\documentclass[lettersize,journal]{IEEEtran}

\usepackage{array}
\usepackage[caption=false,font=normalsize,labelfont=sf,textfont=sf]{subfig}
\usepackage{textcomp}
\usepackage{stfloats}
\usepackage{url}
\usepackage{verbatim}
\usepackage{graphicx}
\usepackage{cite}
\usepackage{amsmath,amsfonts, amssymb, amsthm}  
\theoremstyle{remark}
\newtheorem{remark}{Remark}

\usepackage[T1]{fontenc}    

\usepackage{algorithmicx, algpseudocode, algorithm}  

\usepackage{mathtools, mathcomp}  
\usepackage{siunitx}
\usepackage{comment}
\usepackage{xcolor}
\usepackage{multirow}
\usepackage{enumerate}
\usepackage{mleftright}
\mleftright

\mathtoolsset{showonlyrefs=true} 

\usepackage{flushend}

\newcommand{\Of}[1]{\left({#1}\right)} 
\newcommand{\thetaOf}[2]{\mathrm{\theta}_{#1}\left( {#2} \right)} 
\newcommand{\ThetaOf}[2]{\mathrm{\Theta}_{#1}\left( {#2} \right)} 
\newcommand{\Norm}[2]{\left\lVert {#1} \right\rVert _ {{#2}}} 

\newcommand{\Sequence}[4]{\left\{ {#1}^{#2} \left(k\right) \right\}_{k = {#3}}^{{#4}}}

\newcommand{\BvecC}[4]{\begin{bmatrix} {#1}\left({#2}\right) & {#1}\left({#3}\right) & \cdots & {#1}\left({#4}\right) \\ \end{bmatrix} }
\newcommand{\BvecS}[4]{\begin{bmatrix} {#1}_{{#2}} & {#1}_{{#3}} & \cdots & {#1}_{{#4}} \\ \end{bmatrix} }

\usepackage[]{hyperref}
\hypersetup{%
 setpagesize=false,%
 bookmarks=true,%
 bookmarksdepth=tocdepth,%
 bookmarksnumbered=true,%
 colorlinks=false,%
 pdftitle={},%
 pdfsubject={},%
 pdfauthor={},%
 pdfkeywords={}
 }

\begin{document}

\title{Sparse identification of nonlinear dynamics with library optimization mechanism: Recursive long-term prediction perspective}

\author{Ansei Yonezawa,~\IEEEmembership{Member,~IEEE,}, Heisei Yonezawa,~\IEEEmembership{Member,~IEEE,}, Shuichi Yahagi,~\IEEEmembership{Member,~IEEE,}, Itsuro Kajiwara, Shinya Kijimoto, Hikaru Taniuchi, Kentaro Murakami

\thanks{\textcolor{red}{This is the accepted version of the paper published in \textit{IEEE Transactions on Cybernetics}, doi: \href{https://ieeexplore.ieee.org/document/11365958}{10.1109/TCYB.2026.3652850}.}}
\thanks{This work was supported in part by JSPS KAKENHI Grant Numbers JP23K19084 and JP25K17561, and in part by the funding provided by ISUZU Advanced Engineering Center.}
\thanks{Ansei Yonezawa and Shinya Kijimoto are with the Department of Mechanical Engineering, Kyushu University, Fukuoka 819-0395, Japan (e-mail: \texttt{ayonezawa[at]mech.kyushu-u.ac.jp}).}
\thanks{Heisei Yonezawa and Itsuro Kajiwara are with the Division of Mechanical and Aerospace Engineering, Hokkaido University, Sapporo 060-0808, Japan.}
\thanks{Shuichi Yahagi was with the 6th Research Department, ISUZU Advanced Engineering Center, Ltd., Fujisawa 252-0881, Japan. He is now with the Department of Mechanical Engineering, Tokyo City University, Tokyo 158-8557, Japan.}
\thanks{Hikaru Taniuchi and Kentaro Murakami are with the Division of Human Mechanical Systems and Design, Hokkaido University, Sapporo 060-0808, Japan.}
}

\markboth{Journal of \LaTeX\ Class Files,~Vol.~xx, No.~x, xxxx~20xx}%
{Shell \MakeLowercase{\textit{et al.}}: A Sample Article Using IEEEtran.cls for IEEE Journals}


\maketitle

\begin{abstract}
The sparse identification of nonlinear dynamics (SINDy) approach can discover the governing equations of dynamical systems based on measurement data, where the dynamical model is identified as the sparse linear combination of the given basis functions. A major challenge in SINDy is the design of a library, which is a set of candidate basis functions, as the appropriate library is not trivial for many dynamical systems. To overcome this difficulty, this study proposes SINDy with library optimization mechanism (SINDy-LOM), which is a combination of the sparse regression technique and the novel learning strategy of the library. In the proposed approach, the basis functions are parametrized. The SINDy-LOM approach involves a two-layer optimization architecture: the inner-layer, in which the data-driven model is extracted as the sparse linear combination of the candidate basis functions, and the outer-layer, in which the basis functions are optimized from the viewpoint of the recursive long-term (RLT) prediction accuracy; thus, the library design is reformulated as the optimization of the parametrized basis functions. The dynamical model obtained by SINDy-LOM has good interpretability and usability, as this approach yields a parsimonious closed-form model. The library optimization mechanism significantly reduces user burden. The RLT perspective improves the reliability of the resulting model compared with the traditional SINDy approach that can only ensure the one-step-ahead prediction accuracy. 
The effectiveness of the proposed approach is verified through numerical experiments. 
\end{abstract}


\section{Introduction}  \label{S_Introduction}

\subsection{Motivation}  \label{sS_Motivation}
\IEEEPARstart{A}{dynamical} model, a mathematical description of dynamics of the system, plays a crucial role in order to analyze, predict, and control the behavior of a dynamical system. Modeling according to some basic assumptions, referred to as first-principles modeling (e.g., Newton's laws of motion, conservation of energy) is a traditional approach to obtain a dynamical model of a system at hand. However, the first-principles modeling approach is rarely applicable to real-world systems due to their complexity and uncertainty. Although bold approximations of the working principle of the system may make first-principles modeling feasible, the resultant model suffers from discrepancies between the actual system and the model \cite{Yu2024}, i.e., a modeling error. First-principles models are often too complicated to be used in practical applications such as controller design \cite{Zhang2023}.

Recent advancements in mathematical methods, sensing techniques, and computational resources have enabled us to tackle various complex problems in science and engineering in a data-driven manner \cite{Brunton2022}, such as fault diagnosis \cite{Guo2025,Ren2024}, controller design \cite{Yonezawa2024, Yan2024,Yonezawa2025,Yonezawa2025_TSMC}, medical diagnosis \cite{Zuo2024,Sun2024}, fluid dynamics \cite{Lino2023, Heiland2025}, and robotics \cite{Tan2024, Lian2023, Wang2025}. In the field of dynamical systems, various methodologies have been studied to distill the dynamical model from data, i.e., data-driven modeling, such as physics-guided deep learning \cite{Yu2024}, Gaussian process \cite{Beckers2022,Wu2024,Hashimoto2025}, Koopman operator \cite{Donge2024,Guo2025_Koopman,Kajikawa2023}, genetic programming \cite{Zhang2023}, sparse null-subspace learning \cite{Li2022}, symbolic regression via neural networks \cite{Boddupalli2023}.
Various architectures of recurrent neural networks have been developed for identifying and controlling nonlinear dynamical systems \cite{Kumar2023,Kumar2022,Kumar2023_SC}. Although recurrent neural networks are powerful tools for system identification and control, deriving explicit state-space representations or dynamic properties from the learned networks remains challenging due to their complex internal structures. 

Sparse identification of nonlinear dynamics (SINDy) \cite{Brunton2016,Brunton2025} is a framework for identifying governing equations of dynamical systems from data. The SINDy method seeks a parsimonious model, which is composed of the minimum required terms, via sparse regression. Thus, SINDy has several advantages such as interpretability and generalizability of the resultant model \cite{Champion2019}, ease of incorporating prior knowledge \cite{Cao2022}, feasibility under limited data availability \cite{Kaiser2018}, and robustness to noise \cite{Kaiser2018}. Owing to these advantages, the SINDy approach has attracted considerable attention in data-driven science and engineering. 

Many studies have demonstrated various engineering applications of the SINDy framework. For example, mechatronic systems have been modeled on the basis of SINDy (e.g., robot systems \cite{Bhattacharya2022,Papageorgiou2024, Nguyen2024}, actuators \cite{Lee2025}). SINDy-based data-driven models have been used to control power systems \cite{Cai2023,Khazaei2024}. 
Unifying reinforcement learning and SINDy can provide an interpretable control policy with a reduced number of interactions with a full-order environment model \cite{Zolman2025}.
Several studies have used SINDy for tackling mechanical engineering problems involving complex dynamics, including control of aircraft systems \cite{Cao2022,Cao2025} and engine systems \cite{Yahagi2025}. Owing to the drawbacks of the first-principles approach, SINDy is well suited for modeling and controlling chemical processes \cite{Abdullah2023}. SINDy has also been used to reconstruct mechanical models from undersampled MRI spectral data, which has potential applications in the dynamical modeling of in vivo soft tissue \cite{Heesterbeek2024}. Such a wide variety of successful applications demonstrates the utility and importance of the SINDy framework.

\subsection{Related work (algorithmic progress in SINDy)}  \label{sS_Related_Work}
The original SINDy approach \cite{Brunton2016} identifies the ordinary differential equation (ODE) model as a linear combination of nonlinear functions, which are parsimoniously chosen from the predetermined set of basis functions (the set of basis functions is referred to as the \emph{library}). To broaden its applicability, one study has proposed a technique to handle rational function nonlinearities \cite{Kaheman2020}; several algorithms based on sparse regression have been developed for obtaining partial differential equation (PDE) models from data \cite{Messenger2021}. The SINDy framework can be applied to both autonomous and non-autonomous dynamical systems \cite{Kaiser2018}.

Discovering governing laws using noisy data is a major challenge in data-driven methods including SINDy. Although the SINDy approach has intrinsic noise robustness owing to sparse regression \cite{Brunton2016,Kaiser2018}, many studies have focused on developing SINDy variants with enhanced noise robustness, such as the ensembling technique (E-SINDy) \cite{Fasel2022}, subsampling with co-teaching \cite{Abdullah2023}, and noise estimation \cite{Cao2025}. The parallel implicit SINDy (SINDy-PI) algorithm can discover implicit ODEs and PDEs using noisy data \cite{Kaheman2020}. Noise is especially disruptive when computing the derivative information from numerical differentiations. A variant of SINDy inspired by the fourth-order Runge--Kutta method \cite{Goyal2022} does not require the derivative information explicitly; hence, it has high noise robustness. 

The symbolic regression framework including SINDy can identify the closed-form model. This feature is advantageous not only for interpretability but also in facilitating the incorporation of prior knowledge about the system into both the learning process and the resulting model. For example, Lagrangian-SINDy integrates the knowledge of classical mechanics \cite{Chu2020}. Using polynomial basis functions, one study has formulated the sparse regression problem with side information constraints as a sequence of sum-of-square programming \cite{Machado2024}.  In addition to algorithmic improvements, SINDy can integrate prior knowledge by including knowledge-oriented basis functions in the library (e.g., \cite{Cao2025, Abdullah2023, Lathourakis2024,Burrage2024}).

Notably, the selection of the library plays a critical role in the prediction accuracy and simplicity of the resulting model \cite{Abdullah2023}. This is because the SINDy approach strongly relies on the assumption that the dynamical behavior of the system to be identified can be well represented by the linear combination of a few basis functions in the library. A typical choice for the basis functions is elementary functions that can construct universal approximators (e.g., polynomials, trigonometric functions). Nevertheless, the \emph{sparse} linear combination of elementary functions does not necessarily capture the dynamic behavior accurately. If the prior knowledge of the system is available mathematically, the basis functions may be chosen according to the prior knowledge \cite{Cao2025, Abdullah2023, Lathourakis2024, Burrage2024}. 
However, the exact parameter values of such physics-informed functions are typically unknown.
Moreover, an excessively large library makes the sparse regression ill-conditioned, resulting in the failure of model discovery \cite{Kaheman2020,Kreikemeyer2024}. Thus, in the SINDy framework, the design of the library is not straightforward and it requires special attention. 

To address the challenge of designing the library for SINDy, several studies have proposed automatic library design mechanisms, in which the sparse regression procedure for distilling dynamical models is embedded into optimization schemes. For example, the optimal library for coupled-SINDy \cite{Burrage2024} is discovered as a subsample of the set of basis functions that express the coupling between system components \cite{Kreikemeyer2024}. 
In \cite{Jiang2025}, the basis functions are optimized through evolutionary computation as binary tree structures composed of elementary operators and operands. 
In data-driven model discovery, it is important to consider not only the one-step-ahead prediction error but also the multi-step or long-term prediction error, as discussed in \cite{Stepaniants2024}.
Incorporating the consideration of long-term prediction error into the automatic library design of SINDy enhances the reliability of the identified model. This enhancement facilitates the broader adoption of the SINDy framework across diverse domains, thereby advancing data-driven science and engineering.

\subsection{Contribution and novelty}  \label{sS_Contribution}
This study proposes SINDy with library optimization mechanism (SINDy-LOM). The core idea of the proposed approach is to reformulate the library design problem for SINDy into a two-layer optimization architecture. In the inner-layer, sparse regression is executed using the parametrized library to distill the SINDy model. The parametrized library is a set of the basis functions with tunable parameters. In the outer-layer, the tunable parameters are optimized on the basis of the recursive long-term (RLT) prediction accuracy of the resulting SINDy model. Consequently, the SINDy-LOM approach performs data-driven modeling via sparse regression and improvement of the library simultaneously. Sparse regression realizes the parsimonious data-driven dynamical model in the closed form, contributing toward the interpretability and generalizability of the resulting model. The library optimization strategy provides good basis functions that can capture the nonlinear dynamical behavior. The optimization-based learning approach alleviates the tedious trial-and-error-based library design. As the library is optimized on the basis of the RLT prediction accuracy, the data-driven model provided by the proposed approach achieves reliable prediction, which is indispensable for practical applications such as a prediction model for model predictive control (MPC). The effectiveness of the proposed approach is verified through numerical experiments. 

The contribution and novelty of this study are summarized as follows:
\begin{enumerate}[(1)]
    \item (\emph{Parsimonious model}) The proposed approach identifies the data-driven model as a sparse linear combination of basis functions. Therefore, the resulting model is interpretable and generalizable in the same manner as the traditional SINDy framework. \label{Contribution_Parsimonious_model}
    \item (\emph{Reliability}) Owing to the library learning mechanism, the proposed approach provides good basis functions for capturing the dynamical behavior of the system accurately. The learning criterion is the accuracy of the RLT prediction. Therefore, compared with the conventional SINDy approach that requires a predetermined library and considers only the one-step-ahead prediction error, the proposed approach can discover the reliable data-driven model and achieve accurate RLT prediction. \label{Contribution_Reliability}
    \item (\emph{Reduced user burden}) In the traditional SINDy approach, the basis functions must be manually specified by the user. Finding an appropriate library is costly, time-consuming, and often nontrivial in many applications. By contrast, the proposed approach automatically designs an appropriate library, thereby alleviating the user's burden. \label{Contribution_Burdens}
    \item (\emph{Versatility}) Previous studies focusing on the library design strongly rely on prior knowledge regarding the system to be modeled (e.g., \cite{Cao2025, Abdullah2023, Lathourakis2024, Burrage2024}). Meanwhile, the proposed approach is applicable not only to systems with readily available prior knowledge but also to those where such knowledge is difficult to obtain. Therefore, this study provides a library design strategy for the SINDy method applicable to various dynamical systems. \label{Contribution_Generality}
\end{enumerate}

In summary, this study enables the development of tractable and reliable data-driven models for a wide range of dynamical systems, thereby advancing data-driven approaches to a broad class of problems involving dynamical systems.

\subsection{Structure of the paper}  \label{sS_Structure_of_the_paper}
The remainder of this paper is organized as follows. Section \ref{S_Basic_idea_of_SINDy} outlines the SINDy method. Section \ref{S_Proposed_approach} presents the proposed approach, i.e., SINDy-LOM. Sections \ref{S_Example_SLR_Numerical_example} and \ref{S_Example_Engine_Numerical_example} demonstrate the validity of the proposed approach. Section \ref{S_Discussion} provides discussion, and Section \ref{S_Conclusion} concludes the paper.

\subsection{Notation}  \label{sS_Notation}
The sets of real numbers and strictly positive real numbers are denoted as $\mathbb{R}$ and $\mathbb{R}_{+}$, respectively. The symbol $\mathbb{R}^{n \times m}$ represents the set of $n \times m$-dimensional real matrices (we write $\mathbb{R}^{n \times 1}$ as $\mathbb{R}^{n}$ for simplicity). Let $\mathrm{diag} \Of{v}$ for 
$v = \BvecS{v}{1}{2}{n} \in \mathbb{R}^{1 \times n}$
denote the diagonal matrix having $v_{1}, v_{2},\ldots, v_{n}$ as the diagonal entries. 
For a real vector $x = \BvecS{x}{1}{2}{n} ^{\top} \in \mathbb{R}^{n}$,
$\mathrm{supp}\Of{x} \triangleq \left\{ i \in \left\{1, 2, \ldots, n \right\}: x_{i} \neq 0 \right\}$. 
The cardinality of a set $\Omega$ is denoted as $\mathrm{card} \Of{\Omega}$. For a matrix $M \in \mathbb{R}^{p \times q}$, the $j$-th row of $M$ is denoted as $M_{j, \star} \in \mathbb{R}^{1 \times q}$.

We use the standard definition of the vector norm: $\Norm{x}{p} \triangleq \left( \sum_{i=1}^{n} \left\lvert x_{i}\right\rvert ^{p} \right)^{\frac{1}{p}}$ for $x \in \mathbb{R}^{n}$ and $x \in \mathbb{R}^{1 \times n}$. Moreover, the $\ell_{0}$ pseudo-norm is defined as 
$\Norm{x}{0} \triangleq \mathrm{card} \Of{\mathrm{supp} \Of{x}}$, i.e., $\Norm{x}{0}$ denotes the number of nonzero elements of $x$. Similarly, for an $n \times m$-dimensional matrix $X$, $\Norm{X}{0} \triangleq \sum_{j=1}^{n} \Norm{X_{j,\star}}{0}$.

\section{Basic idea of SINDy}  \label{S_Basic_idea_of_SINDy}
We briefly outline the basic idea of SINDy. Throughout this study, we focus on discrete-time non-autonomous systems for notational simplicity. Extensions to continuous-time and autonomous systems are straightforward. The discrete-time non-autonomous formulation is useful in various engineering applications (e.g., a prediction model for MPC \cite{Bhattacharya2022}).

\subsection{Data-driven model discovery via sparse regression}  \label{sS_Discovery_via_sparse_regression}
Equation \eqref{Eq:x=f(x,w)} describes a nonlinear dynamical system to be identified:
\begin{equation}
    x\Of{k+1} = f\Of{ x\Of{k}, w\Of{k} },
    \label{Eq:x=f(x,w)}
\end{equation}
where $x\Of{k} \in \mathbb{R}^{n}$ and $w\Of{k} \in \mathbb{R}^{m}$ are the state vector and exogenous inputs (e.g., external disturbances, control inputs), respectively. The time-series data $\Sequence{x}{D}{0}{N}$ and $\Sequence{w}{D}{0}{N}$ of the state and exogenous inputs are obtained from the system \eqref{Eq:x=f(x,w)}. Let the data matrices be defined as follows: $X^{D} \triangleq \BvecC{x^{D}}{0}{1}{N-1} \in \mathbb{R}^{n \times N}$, $X^{D+} \triangleq \BvecC{x^{D}}{1}{2}{N} \in \mathbb{R}^{n \times N}$, $W^{D} \triangleq \BvecC{w^{D}}{0}{1}{N-1} \in \mathbb{R}^{m \times N}$.

In the SINDy method, the dynamical behavior of the system \eqref{Eq:x=f(x,w)} is assumed to be well described by the sparse linear combination of the predetermined basis functions 
$\thetaOf{1}{x\Of{k},w\Of{k}}, \thetaOf{2}{x\Of{k},w\Of{k}}, \ldots, \thetaOf{p}{x\Of{k},w\Of{k}}$, 
where $\mathrm{\theta}_{i}:\mathbb{R}^{n}\times\mathbb{R}^{m} \mapsto \mathbb{R}$.
The library $\ThetaOf{}{x\Of{k},w\Of{k}}$ is the set of basis functions and is defined as
\begin{multline}
\ThetaOf{}{x\Of{k},w\Of{k}} \\ 
\triangleq \begin{bmatrix} \thetaOf{1}{x\Of{k},w\Of{k}} & \thetaOf{2}{x\Of{k},w\Of{k}} & \ldots & \thetaOf{p}{x\Of{k},w\Of{k}} \\ \end{bmatrix}.
\label{Eq: Theta(x,w)}
\end{multline}
The SINDy method seeks the sparse coefficient vectors $\xi_{1}, \xi_{2}, \ldots, \xi_{n}$, with $\xi_i \in \mathbb{R}^p$, that satisfy
\begin{equation}
    \left( X^{D+} \right)^{\top} = \ThetaOf{}{X^{D},W^{D}} \varXi 
    \label{Eq:XD_plus=Theta_XD_WDXi}
\end{equation}
\begin{equation}
    \ThetaOf{}{X^{D},W^{D}} \triangleq
    \begin{bmatrix}
        \ThetaOf{}{x^{D}\Of{0},w^{D}\Of{0}} \\
        \ThetaOf{}{x^{D}\Of{1},w^{D}\Of{1}} \\
        \vdots \\
        \ThetaOf{}{x^{D}\Of{N-1},w^{D}\Of{N-1}} \\
    \end{bmatrix}
    \label{Eq:Theta_XD_WD_Def}
\end{equation}
\begin{equation}
    \varXi  \triangleq \BvecS{\xi}{1}{2}{n}.
    \label{Eq:Xi_Def}
\end{equation}
Finally, the data-driven model of the system \eqref{Eq:x=f(x,w)} is identified as follows \cite{Brunton2025,Bhattacharya2022}:
\begin{equation}
    x\Of{k+1} = \left( \ThetaOf{}{x\Of{k}, w\Of{k}} \varXi \right)^{\top}.
    \label{Eq:x(k+1)=Theta(x,w)Xi}
\end{equation}

A typical approach to obtain $\xi_{1}, \xi_{2}, \ldots, \xi_{n}$ is to solve the regularized least squares problem:
\begin{equation}
    \xi_{i} = \underset{v \in \mathbb{R}^{p}}{\operatorname{arg\,min}} \, 
    \Norm{\ThetaOf{}{X^{D},W^{D}}v - \left( X_{i,\star}^{D+} \right)^{\top}}{2} + \mathcal{R}\Of{v}
    \label{Eq:xi_i=argmin(Theta(XD,Wd))}
\end{equation}
where $\mathcal{R}\Of{v}$ is a suitable sparsity-promoting regularizer. In particular, $\gamma\Norm{v}{0}$ is often used as the regularizer \cite{Brunton2016}, where $\gamma \in \mathbb{R}_{+}$ is a constant weight controlling the degree of sparsity promotion. 
The sequentially thresholded least-squares (STLSQ) algorithm proposed in \cite{Brunton2016} and its variants, e.g., \cite{Zhu2019}, can heuristically solve \eqref{Eq:xi_i=argmin(Theta(XD,Wd))} with $\mathcal{R}\Of{v} = \gamma\Norm{v}{0}$. The convergence property of the STLSQ algorithm has been analyzed in \cite{Zhang2019}. Algorithm \ref{Algorithm_STLSQ} presents an overview of the STLSQ algorithm. 
In Algorithm \ref{Algorithm_STLSQ}, $\lambda \in \mathbb{R}_{+}$ is the threshold for the sparsification; $K_{max}$ represents the maximum number of iterations. In addition, $\xi_{i,\Of{j}}$ denotes the $j$-th component of $\xi_{i}$.

\begin{remark}  \label{Remark_Regularizer}
    In \eqref{Eq:xi_i=argmin(Theta(XD,Wd))}, a variety of regularizers can be employed for $\mathcal{R}\Of{v}$. For example, the sparse relaxed regularized regression approach can solve sparse regression problems while enforcing physical constraints \cite{Champion2020}. A recent study compared various sparse regression techniques for SINDy \cite{Kaptanoglu2023}. In addition to thresholding approaches, sparsity can also be promoted through sensitivity-analysis-based techniques \cite{Naozuka2022}.
\end{remark}

\begin{algorithm}[t]
  \caption{STLSQ algorithm}
      \label{Algorithm_STLSQ}
  \begin{algorithmic}[1]
    \Require{$\ThetaOf{}{X^{D},W^{D}}$, $X_{i, \star}^{D+}$, $\lambda$, $K_{max}$}   
    \Ensure{$\xi_{i}$}  
    \State $\xi_{i} \gets \underset{v \in \mathbb{R}^{p}}{\operatorname{arg\,min}} \, 
    \Norm{\ThetaOf{}{X^{D},W^{D}}v - \left( X_{i,\star}^{D+} \right)^{\top}}{2}$
    \State $\mathcal{I} \gets \left\{j \in \left\{1,2,\cdots, p\right\}: \xi_{i,\Of{j}} \geq  \lambda  \right\}$
    \State $ \xi_{i,\Of{j}} \gets \begin{cases} 
                         \xi_{i,\Of{j}} & \text{if $ j \in \mathcal{I}$} \\
                         0  & \text{else}
                         \end{cases}$
    \State $k \gets 1$
    \While{$k \leq K_{max}$}   
    \Statex \% \texttt{Compute $\xi_i$ by solving a least squares problem over the non-zero components.} \%
    \State $\xi_{i} \gets \underset{v \in \mathbb{R}^{p}: \mathrm{supp}\Of{v}\subseteq\mathcal{I}}{\operatorname{arg\,min}} \, 
    \Norm{\ThetaOf{}{X^{D},W^{D}}v - \left( X_{i,\star}^{D+} \right)^{\top}}{2}$
    \Statex \% \texttt{Find components of $\xi_i$ that are larger than $\lambda$.} \%
    \State $\mathcal{I} \gets \left\{j \in \left\{1,2,\cdots, p\right\}: \xi_{i,\Of{j}} \geq  \lambda  \right\}$
    \Statex \% \texttt{Replace small components with $0$.} \%
    \State $ \xi_{i,\Of{j}} \gets \begin{cases} 
                         \xi_{i,\Of{j}} & \text{if $ j \in \mathcal{I}$,} \\
                         0  & \text{else}
                         \end{cases}$
    \State $k \gets k+1$
    \EndWhile
  \end{algorithmic}
\end{algorithm}   

\subsection{Challenges in the SINDy method}  \label{sS_Challenges_in_SINDy}
Although the SINDy method described in Section \ref{sS_Discovery_via_sparse_regression} is a powerful approach for discovering dynamical models from data, there still remain several challenges.

First, the design strategy of the library $\ThetaOf{}{x\Of{k},w\Of{k}}$ is unclear. The selection of the basis functions 
$\thetaOf{1}{x\Of{k},w\Of{k}},\thetaOf{2}{x\Of{k},w\Of{k}}, \ldots, \thetaOf{p}{x\Of{k},w\Of{k}}$ plays a critical role in SINDy, because SINDy assumes that the dynamical behavior of the system can be well captured by a sparse linear combination of these basis functions. In other words, in the SINDy method, the identification problem of nonlinear dynamics is reformulated into the computation of the sparse coefficient vectors $\xi_{1},\xi_{2}, \ldots, \xi_{n}$ with respect to the predetermined basis functions. Although prior knowledge of the system can provide insights into the choice of basis functions, depending on the application (e.g., aircraft system \cite{Cao2025}, chemical reaction \cite{Abdullah2023}, friction \cite{Lathourakis2024}), such domain-specific knowledge is often difficult to obtain in many practical situations. When prior knowledge is not available, we may employ elementary functions capable of serving as universal approximators, such as polynomials \cite{Brunton2025}. However, the dynamical behavior is not necessarily well represented by their \emph{sparse} linear combination. Moreover, it is undesirable to include too many basis functions in the library, since an excessively large library can make the sparse regression problem ill-conditioned \cite{Kaheman2020, Kreikemeyer2024}. Therefore, selecting appropriate basis functions is a challenging task. 

The second challenge is that the sparse regression \eqref{Eq:xi_i=argmin(Theta(XD,Wd))} does not consider the RLT prediction accuracy of the resultant model \eqref{Eq:x(k+1)=Theta(x,w)Xi}. Specifically, the data-driven model \eqref{Eq:x(k+1)=Theta(x,w)Xi} obtained via solving \eqref{Eq:xi_i=argmin(Theta(XD,Wd))} minimizes the one-step-ahead prediction error $J_{os}$:
\begin{equation}
    J_{os} = \sum_{i=1}^{n} \Norm{\check{X}_{i, \star}^{D+} - X_{i, \star}^{D+}}{2}
    \label{Eq:J_OS=SIGMA}
\end{equation}
\begin{equation}
    \check{X}^{D+} \triangleq \BvecC{\check{x}^{D}}{1}{2}{N}
    \label{Eq:X_Check=[x_Check]}
\end{equation}
\begin{equation}
    \check{x}^{D}\Of{k+1} = \left(\ThetaOf{}{x^{D}\Of{k},w^{D}\Of{k}} \varXi \right)^{\top}.
    \label{Eq:x_Check(k+1)=Theta(xk,wk)Xi}
\end{equation}

Equation \eqref{Eq:J_OS=SIGMA} implies that the sparse regression \eqref{Eq:xi_i=argmin(Theta(XD,Wd))} minimizes the error between the true data $x^{D}\Of{k+1}$ and its one-step-ahead prediction $\check{x}^{D}\Of{k+1}$, computed from the true data $x^{D}\Of{k}$ and $w^{D}\Of{k}$ using \eqref{Eq:x_Check(k+1)=Theta(xk,wk)Xi}. However, this formulation cannot maintain the accuracy of the RLT prediction. In other words, the RLT prediction at the $\left( k+\tau \right)$-th time step ($\tau = 2, 3, \ldots$) obtained from $x^{D}\Of{k}$ and $w^{D}\Of{k}$ may be inaccurate due to the propagation of the prediction errors through recurrent use of the model. The inaccuracy of the RLT prediction is problematic in many applications.

\begin{remark} \label{Remark_Innacuraccy_of_RLT_Prediction}
    The inaccuracy of the RLT prediction due to the recurrent use of the model has been discussed in the context of the Koopman operator framework \cite{Lawryczuk2024} and the SINDy framework \cite{Yahagi2025_arXiv}. However, the challenge of designing the library for the SINDy approach was not addressed in \cite{Yahagi2025_arXiv}.
\end{remark}

\section{Proposed approach}  \label{S_Proposed_approach}
This section presents the proposed approach, namely SINDy-LOM, which consists of library parametrization and model discovery via the two-layer optimization architecture. Equation \eqref{Eq:x=f(x,w)} represents the dynamical system to be identified.  

\subsection{Parametrized library}  \label{sS_Parametrized_library}
To address the challenges in library design, we propose the use of the parametrized library. In particular, in the SINDy-LOM strategy, the library contains the basis functions
$\thetaOf{1}{x\Of{k},w\Of{k};\phi_{1}}, \thetaOf{2}{x\Of{k},w\Of{k};\phi_{2}}, \ldots, \thetaOf{p}{x\Of{k},w\Of{k};\phi_{p}}$,
where $\thetaOf{i}{x,w;\phi_{i}}$ implies that the function $\theta_{i}$ has the tunable parameter $\phi_{i} \in \mathbb{R}^{o_{i}}$. 
The parametrized library is expressed as 
\begin{multline}
    \ThetaOf{}{x, w; \varPhi} \\
    \triangleq \begin{bmatrix} \thetaOf{1}{x,w; \phi_{1}} & \thetaOf{2}{x,w; \phi_{2}} & \cdots & \thetaOf{p}{x,w; \phi_{p}} \end{bmatrix}
    \label{Eq: Theta(x,w,Phi)}
\end{multline}
where $\varPhi$ is a parameter vector comprising all elements of $\phi_{1}, \phi_{2}, \ldots, \phi_{p}$. The tuning parameters $\phi_{1}, \phi_{2}, \ldots, \phi_{p}$ are optimized on the basis of the RLT prediction accuracy as discussed in Section \ref{sS_SINDy_LOM}. Optimizing the tuning parameters of the basis functions corresponds to automatically designing the appropriate basis functions capturing the dynamical behavior of the system. Therefore, searching for the values of $\varPhi$ results in learning the appropriate library. 

A variety of functions can be employed for $\thetaOf{i}{x,w;\phi_{i}}$. Elementary functions capable of serving as universal approximators are suitable for constructing the parametrized library. For example, the polynomials 
$\prod_{i,j}^{} \left(x_{i} - \eta_{i} \right)^{\alpha_{i}} \left(w_{j} - \mu_{j} \right)^{\beta_{j}}$ 
with the tunable parameter $\phi = \begin{bmatrix} \eta_{1} & \cdots & \eta_{n} & \mu_{1} & \cdots & \mu_{m} \end{bmatrix}^{\top} $
corresponds to the Taylor approximation around $\phi$. 
The exponents $\alpha_{i}$ and $\beta_{j}$ can be treated either as fixed parameters or as design variables that are tuned by the proposed approach.
Moreover, the library can be partially parametrized: 
$\ThetaOf{}{x, w; \varPhi} = \begin{bmatrix} \theta_{1} & \cdots & \theta_{p_{d}} & \theta_{p_{d}+1} & \ldots & \theta_{p} \end{bmatrix}$,
where $\thetaOf{1}{x, w}, \ldots, \thetaOf{p_{d}}{x, w}$ are fixed and 
$\thetaOf{p_{d}+1}{x, w; \phi_{1}}, \ldots, \thetaOf{p}{x, w; \phi_{p-p_{d}}}$
are parametrized and optimized. 
Basis functions with explicit closed-form expressions, rather than opaque representations, are easier to interpret. Accordingly, to enhance the interpretability of the resulting model, the parametrized library should be composed of elementary functions with well-established mathematical properties (e.g., trigonometric and exponential functions), as well as physics-informed functions derived from prior knowledge.

\subsection{Two-layer optimization architecture (SINDy-LOM)}  \label{sS_SINDy_LOM}
We propose the data-driven discovery technique based on sparse regression and library optimization, namely the SINDy-LOM approach. Here, it is assumed that the datasets $\mathcal{X}^{SR} = \Sequence{x}{SR}{0}{N^{SR}}$, $\mathcal{W}^{SR} = \Sequence{w}{SR}{0}{N^{SR}}$ and the datasets $\mathcal{X}^{LL_{i}} = \Sequence{x}{LL_{i}}{0}{N_{i}^{LL}}$, $\mathcal{W}^{LL_{i}} = \Sequence{w}{LL_{i}}{0}{N_{i}^{LL}}$ ($i = 1, 2, \ldots, M$) are collected, where $N^{SR}$ and $N_{i}^{LL}$ denote the numbers of the sampling points of $\left(\mathcal{X}^{SR}, \mathcal{W}^{SR}\right)$ and $\left(\mathcal{X}^{LL_{i}}, \mathcal{W}^{LL_{i}}\right)$, respectively. Here, $\mathcal{X}^{SR}$ and $\mathcal{W}^{SR}$ are used for sparse regression, whereas the library is learned using $\mathcal{X}^{LL_{i}}$ and $\mathcal{W}^{LL_{i}}$. We cast the library learning and model identification into the following optimization problem:
\begin{equation}
    \varPhi^{\ast} = \underset{\varPhi}{\operatorname{arg\,min}} \, J_{ms}\Of{\varPhi}
    \label{Eq:Phi^star=argmin(Jms)}
\end{equation}
\begin{multline}
    J_{ms}\Of{\varPhi} \\ \triangleq \frac{1}{QR} \sum_{i=1}^{M} q_{i}
        \left\{ \frac{1}{\sqrt{N_{i}^{LL}}} 
                \sum_{j=1}^{n} r_{j} \frac{\Norm{\hat{E}_{j, \star}^{LL_{i}}}{2}}{\Norm{X_{j, \star}^{LL_{i}}}{2}}
        \right\}
        + \kappa \Norm{\mathit{\Xi}}{0}
    \label{Eq:Jms=def}
\end{multline}
where
\begin{equation}
    \hat{E}_{j, \star}^{LL_{i}} \triangleq \hat{X}_{j, \star}^{LL_{i}} - X_{j, \star}^{LL_{i}}
    \label{Eq:Ehat^LLi=def}
\end{equation}
\begin{equation}
    \hat{X}^{LL_{i}} \triangleq \BvecC{\hat{x}^{LL_{i}}}{0}{1}{N_{i}^{LL}-1}
    \label{Eq:Xhat^LLi=def}
\end{equation}
\begin{equation}
    X^{LL_{i}} \triangleq \BvecC{x^{LL_{i}}}{0}{1}{N_{i}^{LL}-1}
    \label{Eq:X^LLi=def}
\end{equation}
\begin{equation}
    \hat{x}^{LL_{i}}\Of{k+1} = \left( \ThetaOf{}{\hat{x}^{LL_{i}}\Of{k},w^{LL_{i}}\Of{k}; \varPhi} \varXi \right)^{\top}  
    \label{Eq:xhat^LLi(k+1)}
\end{equation}
\begin{equation}
    \hat{x}^{LL_{i}}\Of{0} = x^{LL_{i}}\Of{0}
    \label{Eq:xhat^LLi(0)}
\end{equation}  
\begin{equation}
    \varXi = \BvecS{\xi}{1}{2}{n}
    \label{Eq:Xi=[xi_j]}
\end{equation}
\begin{multline}
    \xi_{j} = \underset{v \in \mathbb{R}^{p}}{\operatorname{arg\,min}} \, 
    \Norm{\ThetaOf{}{X^{SR},W^{SR};\varPhi}v - \left( X_{j,\star}^{SR+} \right)^{\top}}{2} \\ + \mathcal{R}\Of{v}
    \label{Eq:xi_i=argmin(Theta(XSR,WSR;Phi))}
\end{multline}
\begin{equation}
    X^{SR} \triangleq \BvecC{x^{SR}}{0}{1}{N^{SR}-1}
    \label{Eq:X^SR=def}
\end{equation}
\begin{equation}
    X^{SR+} \triangleq \BvecC{x^{SR}}{1}{2}{N^{SR}}
    \label{Eq:X^SR+=def}
\end{equation}
\begin{equation}
    W^{SR} \triangleq \BvecC{w^{SR}}{0}{1}{N^{SR}-1}
    \label{Eq:W^SR=def}
\end{equation}
\begin{multline}
    \ThetaOf{}{X^{SR},W^{SR}; \varPhi} \\ \triangleq
    \begin{bmatrix}
        \ThetaOf{}{x^{SR}\Of{0},w^{SR}\Of{0}; \varPhi} \\
        \ThetaOf{}{x^{SR}\Of{1},w^{SR}\Of{1}; \varPhi} \\
        \vdots \\
        \ThetaOf{}{x^{SR}\Of{N^{SR}-1},w^{SR}\Of{N^{SR}-1}; \varPhi} \\
    \end{bmatrix}
    \label{Eq:Theta_XSR_WSR_Def}
\end{multline}
and $\mathcal{R}\Of{v}$ is some suitable sparsity-promoting regularizer. The term $\kappa \Norm{\varXi}{0}$ is a sparsity-promoting term for optimizing the library, where $\kappa \in \mathbb{R}_{+} \cup \left\{ 0  \right\}$; it facilitates the search for the appropriate library yielding the parsimonious model. In \eqref{Eq:Jms=def}, $q_{i}$ and $r_{j}$ are positive constant weights and $Q \triangleq \sum_{i}^{}q_{i}$ and $R \triangleq \sum_{j}^{}r_{j}$. 
The vector $\hat{E}_{j, \star}^{LL_{i}}$ consists of the RLT prediction errors for the $j$-th component of $x$ in the dataset $\mathcal{X}^{LL_{i}}$.
The relative 2-norm error of the RLT prediction for the $j$-th state component in $\mathcal{X}^{LL_{i}}$ is computed as $\frac{\Norm{\hat{E}_{j, \star}^{LL_{i}}}{2}}{\Norm{X_{j, \star}^{LL_{i}}}{2}}$.
We can control the relative importance of the RLT prediction accuracy for each $\mathcal{X}^{LL_{i}}$ by adjusting $q_{i}$, whereas $r_{j}$ specifies the relative importance of the RLT prediction accuracy for the $j$-th component of $x$. Note that $q_{i}$ and $r_{j}$ can be freely chosen by a user according to the design specification of the resulting model.

Finally, the data-driven model is obtained as 
\begin{equation}
    x\Of{k+1} = \left( \ThetaOf{}{x\Of{k},w\Of{k}; \mathit{\Phi}^{\ast}} \varXi^{\ast} \right)^{\top}  
    \label{Eq:x(k+1)=Theta(x,w,Phi^ast)Xi^ast}
\end{equation}
\begin{equation}
    \varXi^{\ast} = \BvecS{\xi^{\ast}}{1}{2}{n}
    \label{Eq:Xi^ast=[xi_j^ast]}
\end{equation}
\begin{multline}
    \xi_{j}^{\ast} = \underset{v \in \mathbb{R}^{p}}{\operatorname{arg\,min}} \, 
    \Norm{\ThetaOf{}{X^{SR},W^{SR};\varPhi^{\ast}}v - \left( X_{j,\star}^{SR+} \right)^{\top}}{2} \\ + \mathcal{R}\Of{v}.
    \label{Eq:xi_i=argmin(Theta(XSR,WSR;Phi^ast))}
\end{multline}

The proposed optimization problem is formulated with a two-layer architecture. The library optimization is performed in the outer-layer problem \eqref{Eq:Phi^star=argmin(Jms)}, while the data-driven model, given by $\varPhi$ and evaluated in \eqref{Eq:Phi^star=argmin(Jms)}, is obtained via sparse regression \eqref{Eq:xi_i=argmin(Theta(XSR,WSR;Phi))}, which constitutes the inner-layer problem. 
The loss function $J_{ms}\Of{\varPhi}$ is constructed on the basis of $\hat{x}^{LL_{i}}\Of{k}$, which is the $k$-step-ahead RLT prediction of $x^{LL_{i}}\Of{k}$. Note that $\hat{x}^{LL_{i}}\Of{k}$ is obtained by using the data-driven model recurrently from the initial condition data $x^{LL_{i}}\Of{0}$, as described in \eqref{Eq:xhat^LLi(k+1)} and \eqref{Eq:xhat^LLi(0)}. Therefore, solving \eqref{Eq:Phi^star=argmin(Jms)} yields an appropriate library with consideration of the RLT prediction accuracy. 
The coefficients $\varXi$ of the basis functions in \eqref{Eq:xhat^LLi(k+1)} are determined by the sparse regression \eqref{Eq:xi_i=argmin(Theta(XSR,WSR;Phi))}. 
The resultant model \eqref{Eq:x(k+1)=Theta(x,w,Phi^ast)Xi^ast} is constructed as the sparse linear combination of the basis functions optimized for accurate RLT prediction. Consequently, the SINDy-LOM approach yields the parsimonious dynamical model that achieves accurate RLT prediction.

\begin{figure*}
    \begin{center}
    \includegraphics[width=18.2cm]{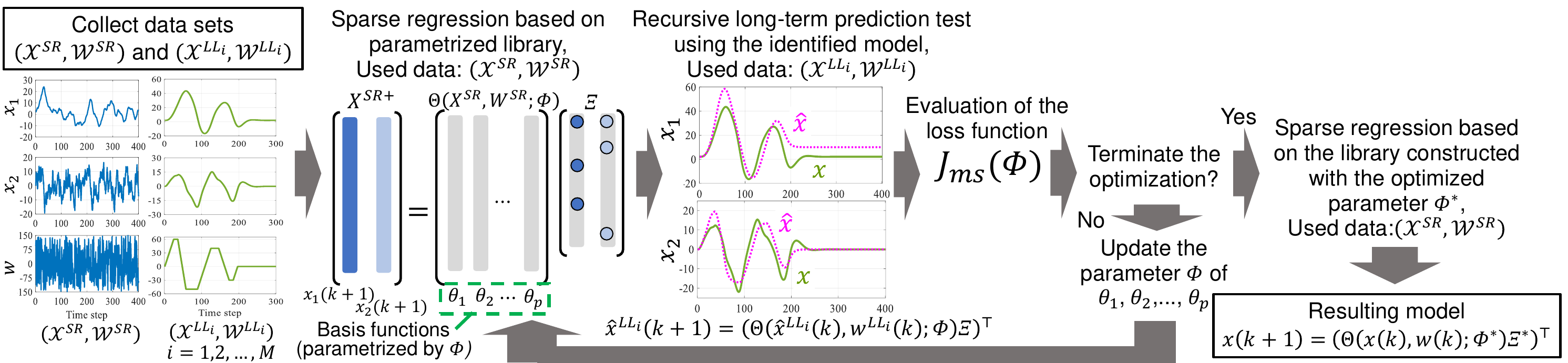}
    \caption{\protect{Schematic of the SINDy-LOM approach.}}
    \label{Fig_Overview_SINDyLOM}
    \end{center}
\end{figure*}

The specific procedure for the SINDy-LOM approach is summarized in Procedure \ref{Procedure_SINDy-LOM}, where the optimization strategy for the outer-layer problem \eqref{Eq:Phi^star=argmin(Jms)} is denoted as OS4OLP. In Procedure \ref{Procedure_SINDy-LOM}, $\mathcal{R}\Of{v} = \gamma \Norm{v}{0}$. 
Fig. \ref{Fig_Overview_SINDyLOM} illustrates the schematic of the SINDy-LOM approach.
Unfortunately, \eqref{Eq:Phi^star=argmin(Jms)} involves a non-convex optimization, which is a drawback of SINDy-LOM. Various optimization techniques such as particle swarm optimization (PSO) and the genetic algorithm (GA) can be employed as the OS4OLP to solve \eqref{Eq:Phi^star=argmin(Jms)}. In the future, we will reformulate \eqref{Eq:Phi^star=argmin(Jms)} to improve the computational efficiency.

Setting the values of $\lambda$ and $\kappa$ has a significant impact on the proposed approach. Larger values enhance model parsimony; however, excessively large values may exclude terms necessary to capture the system dynamics. Conversely, excessively small values can result in overly complex models and potential overfitting. Therefore, $\lambda$ and $\kappa$ should be as large as possible unless the accuracy of the resulting model is compromised.

\begin{algorithm*}
    \renewcommand{\thealgorithm}{1}
    \floatname{algorithm}{Procedure} 

    \% \texttt{Hyperparameters for the sparse regression: $\lambda$, $K_{max}$} \%

    \% \texttt{Hyperparameters for the library optimization: $\kappa$, $q_{i} \, \left(i = 1,\ldots,M\right)$, $r_{j} \, \left(j = 1,\ldots,n\right)$} \%

    \begin{enumerate}
        \item (\emph{Sparse regression with $\varPhi$}) Construct the model $x\Of{k+1} = \left( \ThetaOf{}{x\Of{k},w\Of{k}; \varPhi} \varXi \right)^{\top}$ using $\ThetaOf{}{X^{SR},W^{SR}; \varPhi}$ and $X^{SR+}$ via the STLSQ algorithm;   \label{SINDyLOM_Step_SR}
        \item (\emph{RLT prediction}) Using the model constructed in Step 1, compute $\Sequence{\hat{x}}{LL_{i}}{1}{N_{i}^{LL}}$ for $i = 1,2,\ldots,M$, as shown in \eqref{Eq:xhat^LLi(k+1)} and \eqref{Eq:xhat^LLi(0)};   \label{SINDyLOM_Step_RLT}
        \item (\emph{Loss function evaluation}) Evaluate the value of the loss function $J_{ms}\Of{\varPhi}$;   \label{SINDyLOM_Step_LossEval}
        \item (\emph{Library update}) Update $\varPhi$ according to OS4OLP and the value of $J_{ms}\Of{\varPhi}$ evaluated in Step 3;   \label{SINDyLOM_Step_Library_Update}
        \item (\emph{Library optimization}) Repeat Steps 1--3 until the termination criteria for OS4OLP are satisfied; the optimal value of $\varPhi$ according to Steps 1--3 is denoted as $\varPhi^{\ast}$;   \label{SINDyLOM_Step_Library_Learning}
        \item (\emph{Resulting model}) Derive the resulting data-driven model as \eqref{Eq:x(k+1)=Theta(x,w,Phi^ast)Xi^ast} using the STLSQ algorithm.   \label{SINDyLOM_Step_Result}
    \end{enumerate}
\caption{Pseudocode of the proposed approach (SINDy-LOM)} 
\label{Procedure_SINDy-LOM}
\end{algorithm*}

\begin{remark}  \label{Remark_Difference_from_previous_approach}
    Several studies have proposed library optimization techniques for the parametrized library \cite{Champion2020,OBrien2023,Viknesh2025}. However, unlike SINDy-LOM, previous methods aiming to optimize the parametrized library focus on one-step-ahead prediction accuracy.
\end{remark}

\begin{remark}  \label{Remark_Generality}
    In Steps 1 and 6 in Procedure \ref{Procedure_SINDy-LOM}, various techniques mentioned in Remark \ref{Remark_Regularizer} can replace the STLSQ algorithm to solve the inner problem, depending on the type of $\mathcal{R}\Of{v}$; users can employ various advanced SINDy algorithms (e.g., SINDy-PI, E-SINDy). From this viewpoint, this study provides a general framework to design the library for the SINDy method.
\end{remark}

\begin{remark}   \label{Remark_Side_information}
    A trajectory of a deterministic dynamical system is specified by the initial condition. This fact indicates that the predicted trajectory from the initial condition obtained by recurrently using the identified model must agree with that of the true system. In other words, the behavior of the identified model must coincide with the true dynamical behavior in terms of not only the one-step-ahead prediction but also the RLT prediction. Therefore, the proposed approach incorporates this side information into the data-driven discovery process, thereby improving the reliability of the resulting model.
\end{remark}

\begin{remark}  \label{Remark_Tunable_basis_functions}
    \textcolor{black}{
    In the proposed approach, only the class of basis functions needs to be specified, and their tunable parameters are automatically optimized. In contrast, the standard SINDy approach requires users to define both the class of basis functions and their specific parameter values. From this viewpoint, the proposed approach reduces the design burden for the library compared with traditional SINDy. For example, in many applications, although the class of basis functions is known \emph{a priori}, their parameters are unknown (e.g., chemical processes \cite{Abdullah2023}). The conventional SINDy approach requires careful selection of these parameters, whereas the proposed approach can automatically search for their appropriate values.
    }
\end{remark}

\section{Numerical example \#1}  \label{S_Example_SLR_Numerical_example}

\subsection{Single-link robot system}  \label{sS_Example_SLR_SLR_system}
Equation \eqref{Eq:SLR system} shows the system to be modeled in this example, which is obtained by applying the coordinate transformation $x_{1} \mapsto a_{1}x_{1} + b_{1} $ to the discrete-time single-link robot system given as Example 2 in \cite{Bu2025}:
\begin{equation} 
    \left\{
        \begin{aligned}
            x_{1}\Of{k+1} &= x_{1}\Of{k} + a_{1}h x_{2}\Of{k}  \\
            x_{2}\Of{k+1} &= D_{2} x_{2}\Of{k} + \frac{h}{J} w\Of{k} +D_{s} \sin \Of{\frac{x_{1}\Of{k}}{a_{1}}-\frac{b_{1}}{a_{1}}}
        \end{aligned}
    \right.
    \label{Eq:SLR system}
\end{equation}
where $D_{2} = 1 - \frac{fh}{J}$ and $D_{s} = - \left( 0.5m_{L} + m_{SL} \right)\frac{ghl}{J}$. Here, $m_{L}$, $m_{SL}$, $g$, $l$, $J$, and $f$ denote the mass of the load, the mass of the single link, the gravitational acceleration, the length of the link, the inertia, and the damping, respectively. The sampling time interval is denoted as $h$. In particular, we set $m_{L} = \qty[]{2}{\kilo\gram}$, $m_{SL} = \qty[]{4}{\kilo\gram}$, $g = \qty[]{9.8}{\meter\per\second^{2}}$, $l = \qty[]{0.5}{\meter}$, $J = \qty[]{0.212}{\kilo\gram\cdot\meter^{2}}$, $f = \qty[]{3}{\kilo\gram\cdot\meter^{2}\per\second}$, and $h = \qty[]{0.01}{\second}$. Regarding the coordinate transformation, we set $a_{1} = 10$ and $b_{1} = \frac{2}{3}\pi$.

\subsection{Basis functions}  \label{sS_Example_SLR_Basis_functions}
The basis functions used in this example are summarized in \eqref{Eq:theta_1_22_SLR} and \eqref{Eq:theta_23_25_SLR}: 
\begin{equation}
    \begin{gathered}
    \mathrm{\theta}_{1} = 1,  \;
    \mathrm{\theta}_{2} = x_{1},  \;
    \mathrm{\theta}_{3} = x_{2},  \;
    \mathrm{\theta}_{4} = w,  \; 
    \mathrm{\theta}_{5} = x_{1}^2,  \; \\
    \mathrm{\theta}_{6} = x_{1}x_{2},  \; 
    \mathrm{\theta}_{7} = x_{1}w,  \; 
    \mathrm{\theta}_{8} = x_{2}^{2},  \; 
    \mathrm{\theta}_{9} = x_{2}w,  \;
    \mathrm{\theta}_{10} = w^{2},  \; \\
    \mathrm{\theta}_{11} = \sin\left(x_{1}\right),  \; 
    \mathrm{\theta}_{12} = \sin\left(x_{2}\right),  \; 
    \mathrm{\theta}_{13} = \sin\left(w    \right),  \; \\
    \mathrm{\theta}_{14} = \cos\left(x_{1}\right),  \;
    \mathrm{\theta}_{15} = \cos\left(x_{2}\right),  \; 
    \mathrm{\theta}_{16} = \cos\left(w    \right),  \; \\
    \mathrm{\theta}_{17} = \sin\left(2x_{1}\right),  \; 
    \mathrm{\theta}_{18} = \sin\left(2x_{2}\right),  \; 
    \mathrm{\theta}_{19} = \sin\left(2w    \right),  \; \\
    \mathrm{\theta}_{20} = \cos\left(2x_{1}\right),  \;
    \mathrm{\theta}_{21} = \cos\left(2x_{2}\right),  \; 
    \mathrm{\theta}_{22} = \cos\left(2w    \right),  \; \\
    \end{gathered}
    \label{Eq:theta_1_22_SLR}
\end{equation}
\begin{equation}
    \begin{gathered}
    \mathrm{\theta}_{23} = \sin\left(\nu x_{1} + \psi\right),  \; 
    \mathrm{\theta}_{24} = \sin\left(\nu x_{2} + \psi\right),  \; 
    \mathrm{\theta}_{25} = \sin\left(\nu w + \psi    \right).  \; 
    \end{gathered}
    \label{Eq:theta_23_25_SLR}
\end{equation}
Note that $\mathrm{\theta}_{1}, \ldots, \mathrm{\theta}_{22}$ listed in \eqref{Eq:theta_1_22_SLR} are fixed, whereas $\mathrm{\theta}_{23},\mathrm{\theta}_{24},\mathrm{\theta}_{25}$ shown in \eqref{Eq:theta_23_25_SLR} have tunable parameters $\varPhi = \begin{bmatrix} \nu & \psi \end{bmatrix}^\top$. The ground-truth system \eqref{Eq:SLR system}, with model parameters specified in Section \ref{sS_Example_SLR_SLR_system}, can be expressed as
\begin{equation} 
    \left\{
        \begin{aligned}
            x_{1}\Of{k+1} &= \mathrm{\theta}_{2}\Of{k} + 0.1 \mathrm{\theta}_{3}\Of{k}  \\
             x_{2}\Of{k+1} &= 0.8585 \mathrm{\theta}_{3}\Of{k} + 0.04717 \mathrm{\theta}_{4}\Of{k} -1.1557 \mathrm{\theta}_{23}\Of{k}
        \end{aligned}
    \right.
    \label{Eq:SLR system_theta}
\end{equation}
where $\nu = 0.1$ and $\psi = -0.2094$.

\subsection{Verification setting}  \label{sS_Example_SLR_Verification_setting}
\begin{figure}
    \begin{center}
    \includegraphics[width=8.9cm]{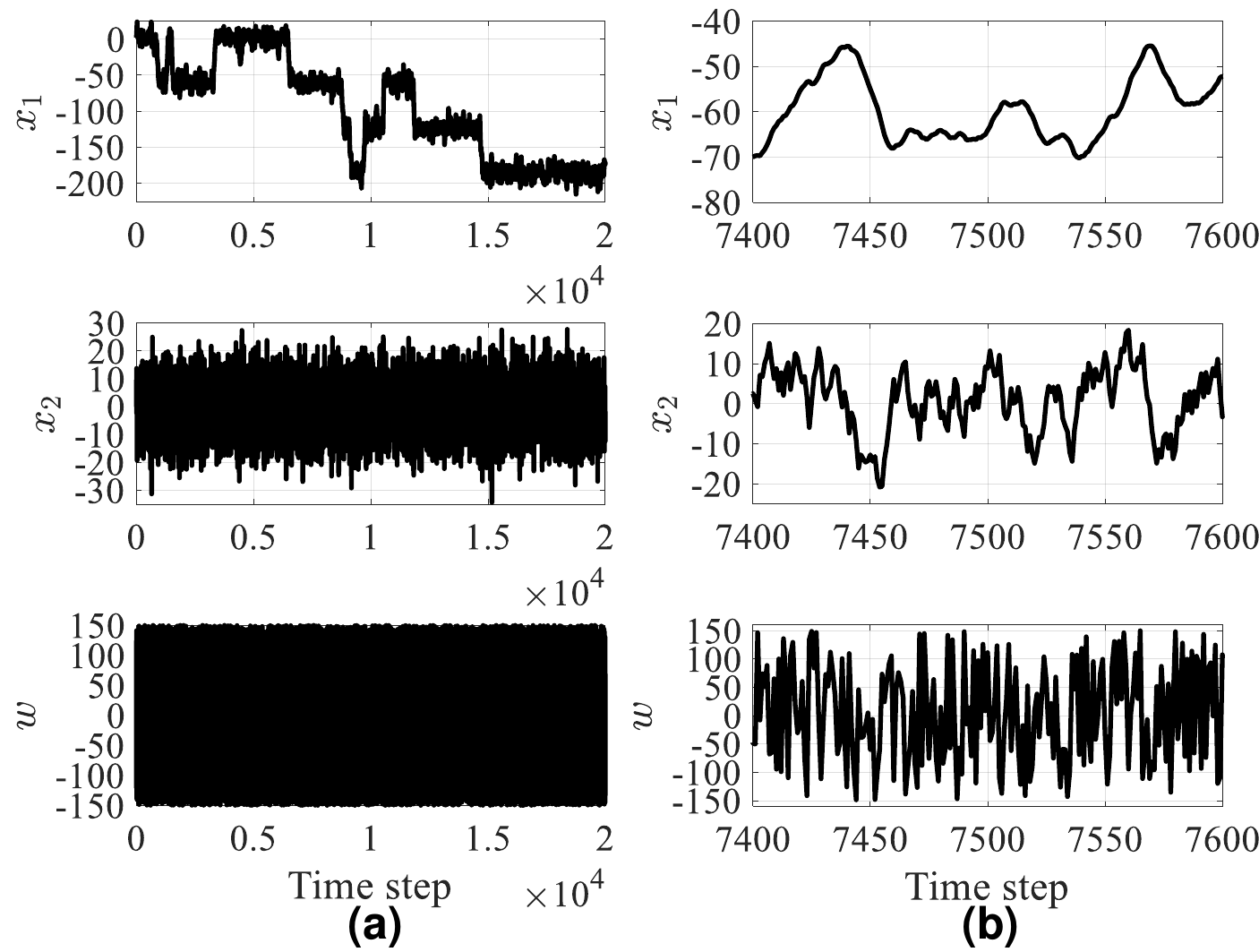}
    \caption{\protect{SR data for Example \#1: (a) overview, and (b) enlarged plot showing the typical behavior in detail. This data is used for both $\left(\mathcal{X}^{SR},\mathcal{W}^{SR} \right)$ and $\left(\mathcal{X}^{LL_{1}},\mathcal{W}^{LL_{1}} \right)$.}}
    \label{Fig_SLR_Data_SR}
    \end{center}
\end{figure}  

\begin{figure}
    \begin{center}
    \includegraphics[width=6.4cm]{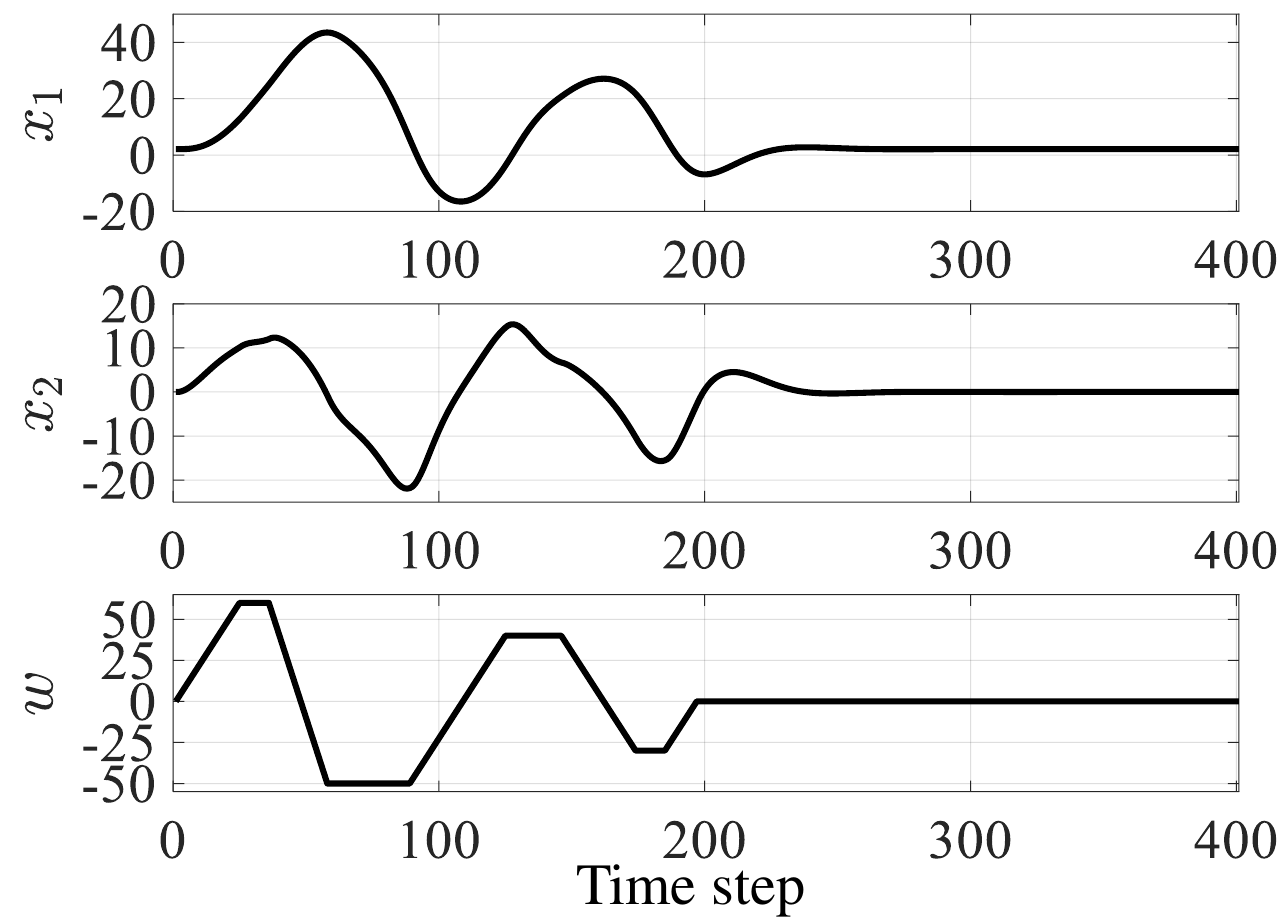}
    \caption{\protect{Overview of the oLL data for Example \#1. This data is used as $\left(\mathcal{X}^{LL_{2}},\mathcal{W}^{LL_{2}} \right)$.}}
    \label{Fig_SLR_Data_oLL}
    \end{center}
\end{figure}  

The proposed approach is compared with the traditional SINDy method. The details of the modeling strategies are summarized in Table \ref{Table_SLR_Overview_of_modeling_strategies}. 
Strategies \#1 and \#2 use the traditional SINDy method with fixed libraries. 
For Strategy \#2, $\varPhi$ is set as $\nu = 18.68119$ and $\psi = 0.29677$, which are randomly sampled from the interval of $\left[-20, 20\right] \times \left[-\pi, \pi\right]$.
In contrast, Strategy \#3 uses SINDy-LOM, where $\varPhi$ is optimized via the library optimization mechanism of SINDy-LOM.
The optimization problem \eqref{Eq:Phi^star=argmin(Jms)} is addressed using the PSO algorithm implemented with MATLAB R2023a's \texttt{particleswarm} function, where the search domain covers $\left[-20, 20\right] \times \left[-\pi, \pi\right]$.
The sparsification parameter $\lambda$ in the STLSQ algorithm is set to $3.5 \times 10^{-2}$ for Strategies \#1--\#3. For Strategy \#3, $\kappa$ is set to $1.0 \times 10^{-3}$. In \eqref{Eq:Jms=def}, we set $q_{1} = q_{2} = 1$ and $r_{1} = r_{2} = 1$.

\begin{table}[]
    \centering
    \caption{Overview of modeling strategies for Example \#1.}
    {
        \tabcolsep = 1.5pt   
\begin{tabular}{c|c|c|c}
\hline
Strategy & Algorithm & Library & $\varPhi$ \\ \hline
\#1 & \begin{tabular}{c} SINDy \\ (Conventional) \end{tabular} & \begin{tabular}{c}$\ThetaOf{1}{x\Of{k},w\Of{k}}$ \\ $=\BvecS{\mathrm{\theta}}{1}{2}{22}$\end{tabular} & -- \\ \hline
\#2 & \begin{tabular}{c} SINDy \\ (Conventional) \end{tabular} & \begin{tabular}{c}$\ThetaOf{2}{x\Of{k},w\Of{k}; \varPhi}$ \\ $=\BvecS{\mathrm{\theta}}{1}{2}{25}$\end{tabular} & \begin{tabular}{c} Fixed, \\ randomly chosen \end{tabular} \\ \hline
\#3 & \begin{tabular}{c} SINDy-LOM \\ (Proposed) \end{tabular} & \begin{tabular}{c}$\ThetaOf{3}{x\Of{k},w\Of{k}; \varPhi}$ \\ $=\BvecS{\mathrm{\theta}}{1}{2}{25}$\end{tabular} & \begin{tabular}{c} Optimized during \\ the modeling process \end{tabular} \\ \hline
\end{tabular}
    }
\label{Table_SLR_Overview_of_modeling_strategies}
\end{table}

The time-series data used for modeling are presented in Figs. \ref{Fig_SLR_Data_SR} and \ref{Fig_SLR_Data_oLL}, hereafter referred to as the \emph{SR data} and the \emph{oLL data}, respectively. 
The SR data is generated by exciting the system with a uniformly distributed random input $w \in \left[-150, 150\right]$. 
Sparse regression for Strategies \#1--\#3 is performed using the SR data. 
In Strategy \#3, the RLT prediction accuracy is assessed with both the SR and oLL datasets. Specifically, the SR data serves for $\left(\mathcal{X}^{SR},\mathcal{W}^{SR} \right)$ and $\left(\mathcal{X}^{LL_{1}},\mathcal{W}^{LL_{1}} \right)$, whereas the oLL data is employed as $\left(\mathcal{X}^{LL_{2}},\mathcal{W}^{LL_{2}} \right)$. Here, $N^{SR} = N_{1}^{LL} = 2.0 \times 10^{4}$ and $N_{2}^{LL} = 4.0 \times 10^{2}$. 
The SR data is designed to uniformly excite the system to capture its dynamic characteristics thoroughly, because it is used for both sparse regression and RLT evaluation. In contrast, the oLL data represents a typical operating scenario. By using both the SR and oLL data to evaluate the RLT performance, we aimed to improve the model's behavior more effectively in practical scenarios.

\begin{remark}  \label{Remark_Implication_of_comparison}
    Strategies \#2 and \#3 have extra basis functions in addition to those used in Strategy \#1. However, these additional basis functions are randomly selected in Strategy \#2, whereas Strategy \#3 optimizes them with respect to the RLT prediction accuracy. Thus, by comparing Strategies \#1 and \#2, we can determine whether augmenting the library is consistently effective in the SINDy approach. The improvement in prediction accuracy achieved by Strategy \#3 confirms the necessity of the proposed library optimization mechanism.
\end{remark}

\subsection{Results}  \label{sS_Example_SLR_Results}
\begin{figure}
    \begin{center}
    \includegraphics[width=8.4cm]{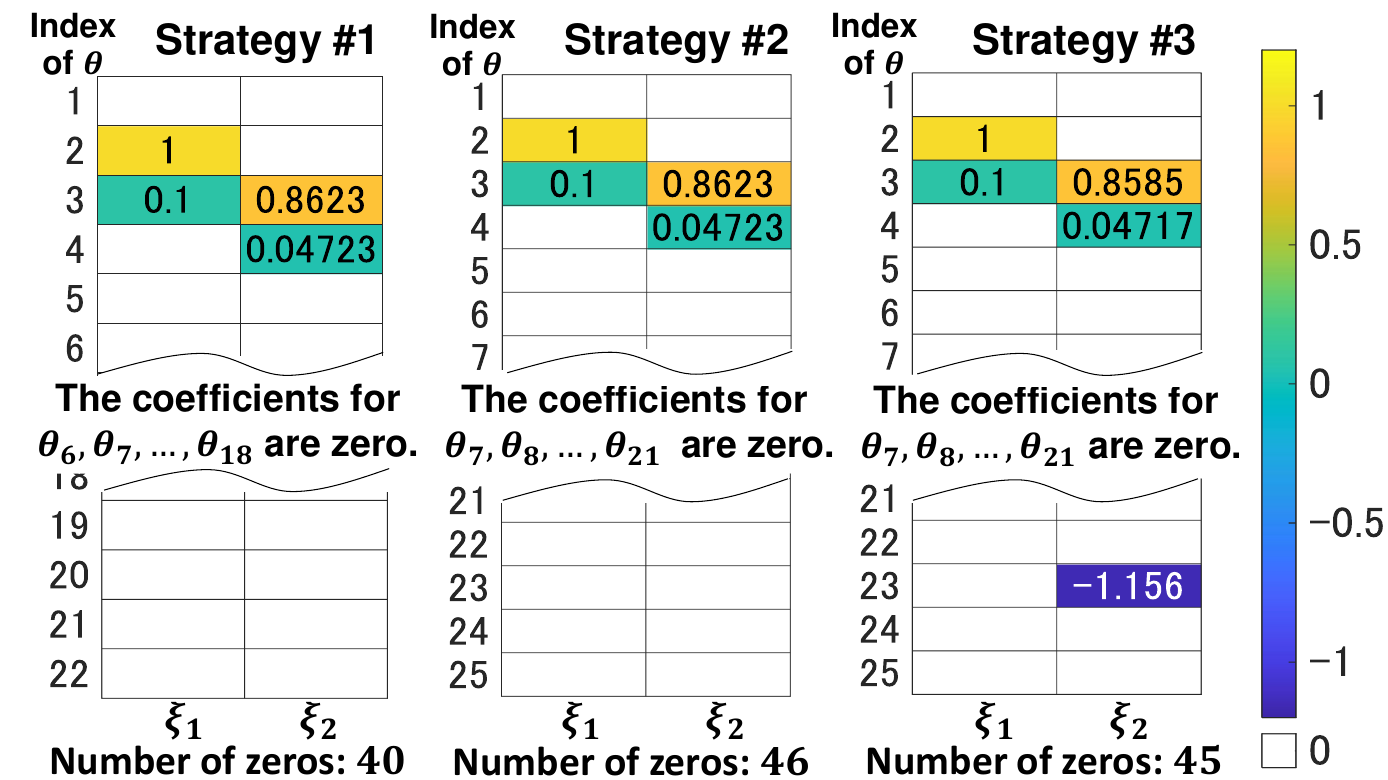}
    \caption{\protect{Visualization of the resulting coefficient matrix $\varXi^{\ast} = \begin{bmatrix} \xi_{1}^{\ast} & \xi_{2}^{\ast} \end{bmatrix}$ obtained by Strategies \#1--\#3 (Example \#1).}}
    \label{Fig_SLR_Xi_Coefficients}
    \end{center}
\end{figure}

Fig. \ref{Fig_SLR_Xi_Coefficients} presents the coefficient matrices $\varXi^{\ast} = \begin{bmatrix} \xi_{1}^{\ast} & \xi_{2}^{\ast} \end{bmatrix}$ obtained from Strategies \#1--\#3. In Fig. \ref{Fig_SLR_Xi_Coefficients}, each row index corresponds to the basis function, and white boxes indicate zero elements. 
The optimized design variables of the library for Strategy \#3, denoted as $\varPhi^{\ast}$, are $\varPhi^{\ast} = \begin{bmatrix} \nu & \psi \end{bmatrix}^\top = \begin{bmatrix} 0.1000 & -0.2094 \end{bmatrix}^\top$.
Fig. \ref{Fig_SLR_Xi_Coefficients} together with $\varPhi^{\ast}$ confirms that Strategy \#3 identifies the ground-truth dynamics \eqref{Eq:SLR system_theta}.

\begin{figure}
    \begin{center}
    \includegraphics[width=8.9cm]{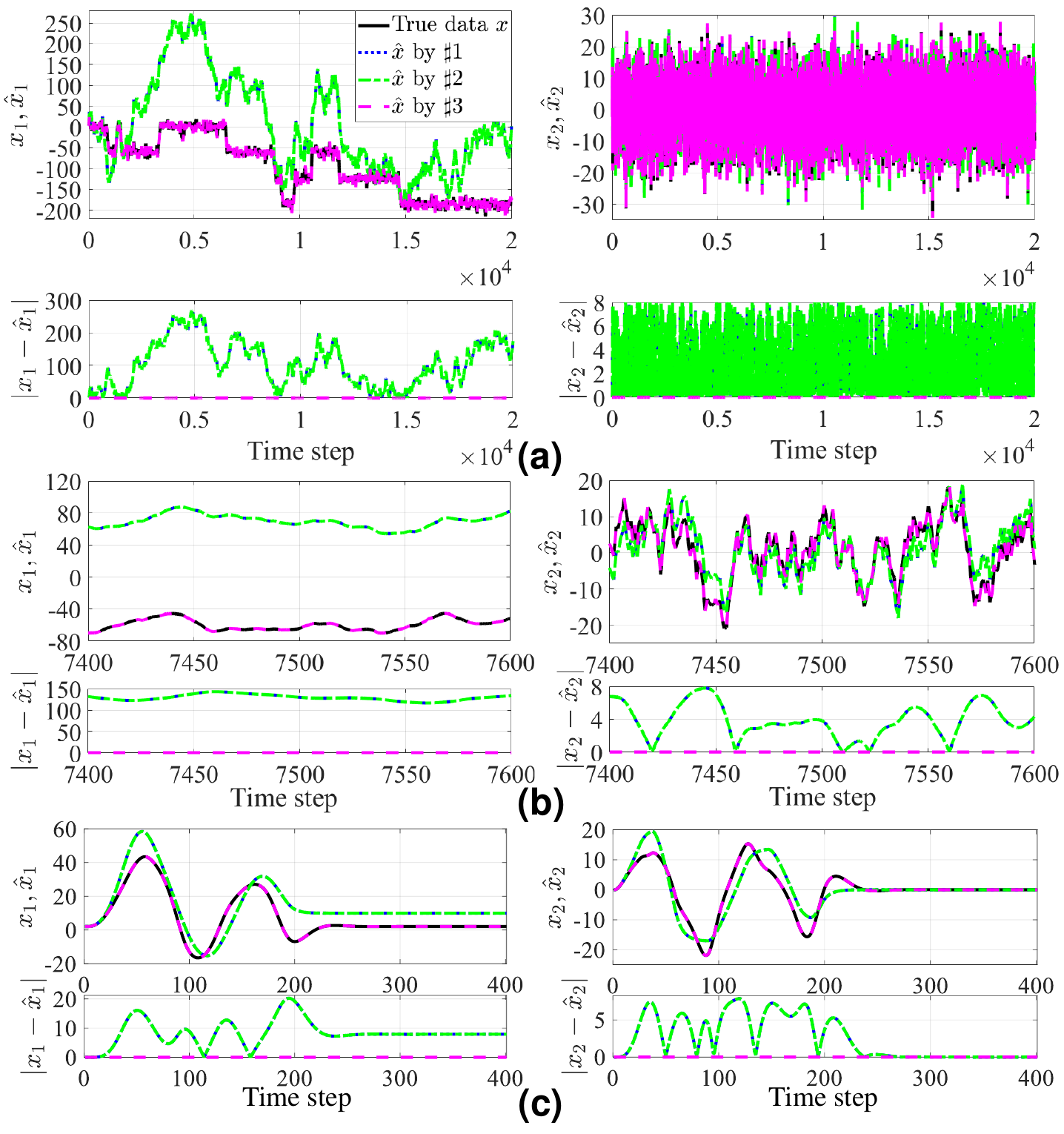}
    \caption{\protect\input{Fig_SLR_RLT_SR_oLL_Caption.tex}}
    \label{Fig_SLR_RLT_SR_oLL}
    \end{center}
\end{figure}

The RLT prediction results for the SR and oLL datasets, are presented in Fig. \ref{Fig_SLR_RLT_SR_oLL}, where the RLT prediction is obtained using the same procedure as in \eqref{Eq:xhat^LLi(k+1)} and \eqref{Eq:xhat^LLi(0)}. Hereafter, consistent with \eqref{Eq:xhat^LLi(k+1)}, $x\Of{k}$ and $\hat{x}\Of{k}$ denote the true state and its RLT prediction, respectively. In Fig. \ref{Fig_SLR_RLT_SR_oLL}, the black solid line represents the true SR or oLL data $x\Of{k}$, while the blue dotted, green dash-dotted, and magenta dashed lines correspond to the RLT predictions $\hat{x}\Of{k}$ obtained from the models identified using Strategies \#1--\#3, respectively.
As shown in Fig. \ref{Fig_SLR_RLT_SR_oLL}, Strategy \#3 achieved a significantly higher RLT prediction accuracy than Strategies \#1 and \#2.
The one-step-ahead prediction results for Strategies \#1--\#3 are shown in Appendix \ref{sS_Appendix_One_step_prediction}. As seen from Fig. \ref{Fig_SLR_RLT_SR_oLL} and the results presented in the Appendix \ref{sS_Appendix_One_step_prediction}, there is a clear discrepancy between the one-step-ahead and RLT predictions.

\subsection{Modeling using noisy data}  \label{sS_Example_SLR_Noisy}
\begin{table*}[]
    \centering
    \caption{Summary of the identification results of the proposed approach using noisy training data. The RLT prediction errors are computed by comparing the RLT prediction results with the true (i.e., noiseless) trajectories.}
    {
    \tabcolsep = 1.5pt   
\begin{tabular}{c|c|cccc}
\hline
\multirow{4}{*}{Noise} & \multirow{4}{*}{\raisebox{8pt}{Identified model}} & \multicolumn{4}{c}{\multirow{2}{*}{\raisebox{8pt}{Error of RLT prediction}}} \\
                  &  & \multicolumn{4}{c}{$
    10^{3} \times \sqrt{\sum_{k} \left\lvert \hat{x}_{i}\Of{k} - x_{i}\Of{k}\right\rvert^{2}} 
    /
    \sqrt{\sum_{k} \left\lvert x_{i}\Of{k}\right\rvert^{2}}
$
}  \\ \cline{3-6} 
$\mathcal{N}\left(0,\sigma_{x}^{2}\right)$ & $ 
\left[
    \mathrm{Truth: \,}
\begin{aligned}
    & x_{1}\Of{k+1} = x_{1}\Of{k} + 0.1 x_{2}\Of{k} \\
    & x_{2}\Of{k+1} = 0.8585 x_{2}\Of{k} + 0.04717 w\Of{k} - 1.1557 \sin \Of{0.1 x_{1}\Of{k} - 0.2094}
\end{aligned}
\right]
$ & \multicolumn{2}{c|}{SR} & \multicolumn{2}{c}{oLL}    \\ \cline{3-6} 
                  &                   & \multicolumn{1}{c|}{\phantom{xxxx}$x_{1}$\phantom{xxx}} & \multicolumn{1}{c|}{\phantom{xxxx}$x_{2}$\phantom{xxx}} & \multicolumn{1}{c|}{\phantom{xxxx}$x_{1}$\phantom{xxx}} & \phantom{xxxx}$x_{2}$\phantom{xxx} \\ \hline
$\sigma_{x} = 0.1$&$
    \begin{aligned}
        x_{1}\Of{k+1} &= 1.0000 x_{1}\Of{k} + 0.1000 x_{2}\Of{k}  \\
        x_{2}\Of{k+1} &= 0.8584 x_{2}\Of{k} + 0.04717 w\Of{k} - 1.1556 \sin\Of{0.1000 x_{1}\Of{k} - 0.2095 }
    \end{aligned}
$
& \multicolumn{1}{c|}{$0.49602$} & \multicolumn{1}{c|}{$1.3900$} & \multicolumn{1}{c|}{$0.51333$} & \multicolumn{1}{c}{$0.73381$}\\ \hline
$\sigma_{x} = 0.2$&$
    \begin{aligned}
        x_{1}\Of{k+1} &= 1.0000 x_{1}\Of{k} + 0.1000 x_{2}\Of{k}  \\
        x_{2}\Of{k+1} &= 0.8580 x_{2}\Of{k} + 0.04719 w\Of{k} - 1.1518 \sin\Of{0.1004 x_{1}\Of{k} - 0.2018}
    \end{aligned}
$
& \multicolumn{1}{c|}{$3.8679$} & \multicolumn{1}{c|}{$12.630$} & \multicolumn{1}{c|}{$6.8333$} & \multicolumn{1}{c}{$7.7092$} \\ \hline
$\sigma_{x} = 0.3$&$
    \begin{aligned}
        x_{1}\Of{k+1} &= 1.0000 x_{1}\Of{k} + 0.0999 x_{2}\Of{k}  \\
        x_{2}\Of{k+1} &= 0.8574 x_{2}\Of{k} + 0.04719 w\Of{k} - 1.1482 \sin\Of{0.1005 x_{1}\Of{k} -0.2112}
    \end{aligned}
$
& \multicolumn{1}{c|}{$6.1427$} & \multicolumn{1}{c|}{$18.578$} & \multicolumn{1}{c|}{$5.4192$} & \multicolumn{1}{c}{$7.0882$} \\ \hline
$\sigma_{x} = 0.4$&$
    \begin{aligned}
        x_{1}\Of{k+1} &= 1.0000 x_{1}\Of{k} + 0.0999 x_{2}\Of{k}  \\
        x_{2}\Of{k+1} &= 0.8566 x_{2}\Of{k}  + 0.04720 w\Of{k} - 1.1385 \sin\Of{0.1008 x_{1}\Of{k} - 0.2131}
    \end{aligned}
$
& \multicolumn{1}{c|}{$10.365$} & \multicolumn{1}{c|}{$34.658$} & \multicolumn{1}{c|}{$9.3112$} & \multicolumn{1}{c}{$12.754$} \\ \hline
$\sigma_{x} = 0.5$&$
    \begin{aligned}
        x_{1}\Of{k+1} &= 1.0000 x_{1}\Of{k} + 0.0996 x_{2}\Of{k} - 0.0473 \sin\Of{0.1018 x_{1}\Of{k} - 0.2254} \\
        x_{2}\Of{k+1} &= 0.8557 x_{1}\Of{k} + 0.04721 x_{2}\Of{k} -1.0781 \sin\Of{0.1018 x_{1}\Of{k} - 0.2254}
    \end{aligned}
$
& \multicolumn{1}{c|}{$19.783$} & \multicolumn{1}{c|}{$56.491$} & \multicolumn{1}{c|}{$31.943$} & \multicolumn{1}{c}{$41.574$} \\ \hline
\end{tabular}
    }
    \label{Table_SLR_Noisy_Result}
\end{table*}


This section demonstrates the effectiveness of the proposed approach under noisy conditions by applying it to noisy datasets.
These datasets are generated by adding zero-mean Gaussian noise $\mathcal{N}\Of{0,\sigma_{x}^{2}}$ to the SR and oLL data depicted in Figs. \ref{Fig_SLR_Data_SR} and \ref{Fig_SLR_Data_oLL} for time steps $k \geq 1$, where $\sigma_{x}^{2}$ denotes the variance of the noise. The hyperparameter values are the same as those used in Section \ref{sS_Example_SLR_Results}.

\begin{figure}
    \begin{center}
    \includegraphics[width=8.95cm]{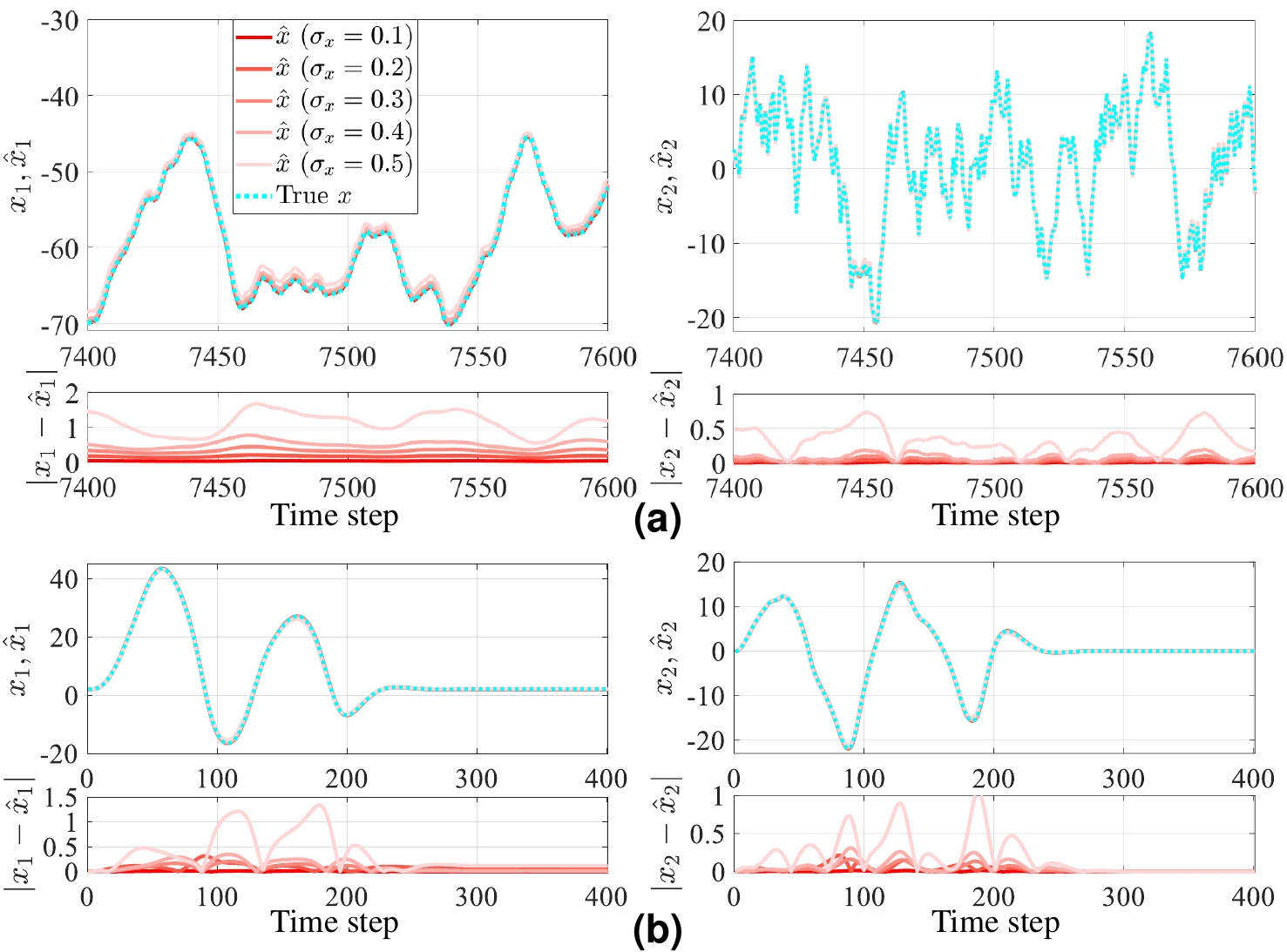}
    \caption{\protect{RLT prediction results and errors obtained from the models identified by the proposed approach using noisy data (Example \#1): (a) an enlarged view of the SR trajectory prediction over the time interval $7400$--$7600$ steps, and (b) the prediction for the oLL trajectory. The models identified from noisy training data demonstrated high RLT prediction accuracy. }}
    \label{Fig_SLR_Noisy_RLT_SR_oLL}
    \end{center}
\end{figure}

The models identified by the proposed approach under various noise levels are summarized in Table \ref{Table_SLR_Noisy_Result}. The RLT prediction results of these models are presented in Fig. \ref{Fig_SLR_Noisy_RLT_SR_oLL}, where the cyan dotted line represents the trajectories of the ground-truth system \eqref{Eq:SLR system_theta}, and the red solid lines represent the prediction results of the identified models. A lighter-colored line corresponds to a model obtained under a higher noise condition. 
Table \ref{Table_SLR_Noisy_Result} and Fig. \ref{Fig_SLR_Noisy_RLT_SR_oLL} demonstrate that the proposed approach is capable of identifying reliable models that maintain high RLT prediction accuracy even under noisy training conditions.

\section{Numerical example \#2}  \label{S_Example_Engine_Numerical_example}

\subsection{Diesel engine airpath system}  \label{sS_Example_Engine_Airpath_system}
\begin{figure}
    \begin{center}
    \includegraphics[width=5.8cm]{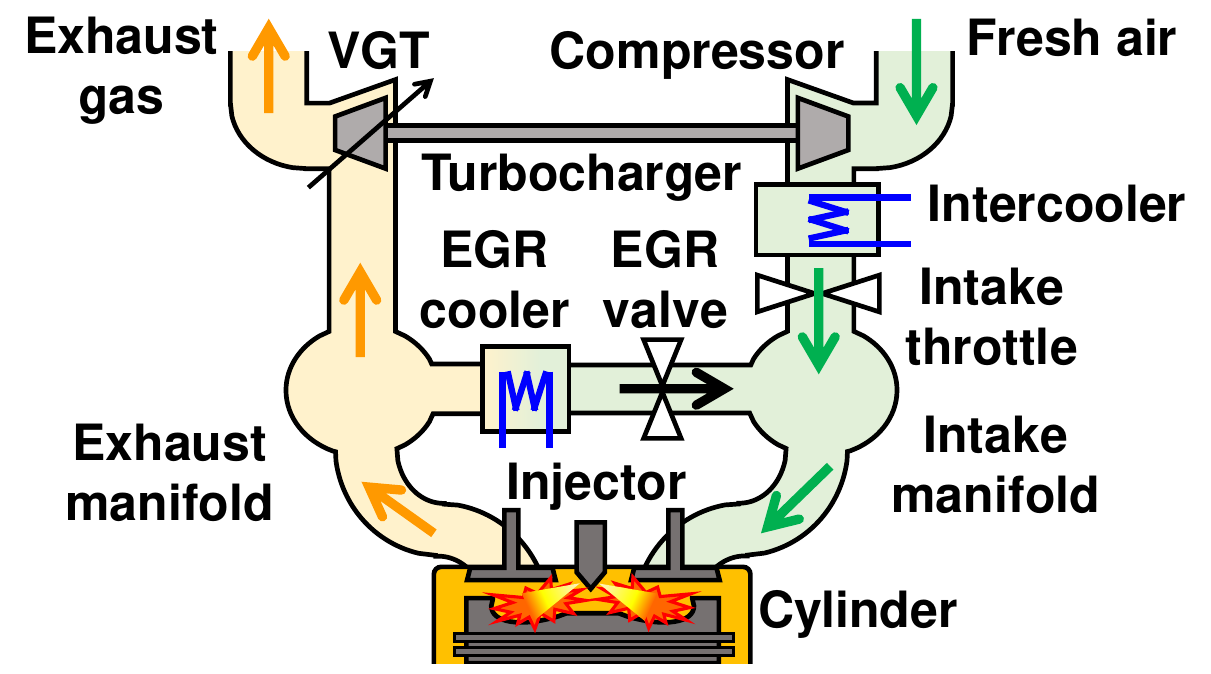}
    \caption{\protect{System to be modeled in Example \#2: diesel engine airpath system.}}
    \label{Fig_Diesel_engine}
    \end{center}
\end{figure}

Fig. \ref{Fig_Diesel_engine} shows the system to be modeled in this example, which is a four-cylinder diesel engine. This engine has an exhaust gas recirculation (EGR) circuit and a variable-geometry turbocharger (VGT). The EGR system introduces the exhaust gas into the intake manifold to maintain the oxygen concentration in the cylinder. The VGT adjusts the pressure inside the intake manifold. In Fig. \ref{Fig_Diesel_engine}, the green and orange arrows illustrate the flows of the intake and exhaust gases, respectively, whereas the black arrow indicates that a part of the exhaust gas is returned to the intake manifold. 
The detailed system descriptions including governing equations can be found in \cite{Yahagi2025,Yahagi2025_arXiv,Hirata2019}.

Table \ref{Table_Variables_of_model} summarizes the variables of the model to be identified. The proposed approach, SINDy-LOM, is used for capturing the dynamical behavior of the intake manifold pressure and the EGR rate, where the exogenous inputs are the engine speed, fuel injection amount, EGR valve opening, and VGT vane closing. The EGR rate is defined as $\frac{100 \times W_{EGR}}{W_{EGR}+W_{PT}} \%$, where $W_{EGR}$ [$\unit[]{\kilogram/\second}$] and $W_{PT}$ [$\unit[]{\kilogram/\second}$] are the EGR flow and the intake throttle flow, respectively. The diesel engine airpath system exhibits strongly nonlinear and coupled dynamics. Therefore, the identification of this system is challenging.

\begin{remark}  \label{Remark_Challenges_engine}
    The nitrogen oxide emission is governed by the in-cylinder oxygen concentration; hence, the EGR rate must be controlled to maintain it. The intake manifold pressure critically affects fuel economy and engine performance. Therefore, accurate modeling of the intake manifold pressure and EGR rate dynamics is essential for diesel engine control. However, the physical phenomena within the diesel engine air-path system are highly complex, thereby making the derivation of a first-principles model challenging. Even if it were constructed, such a model would be too complex for practical use. Therefore, the data-driven modeling problem addressed in this numerical example is of significant practical importance.
\end{remark}

\begin{table}[]
    \centering
    \caption{Variables of the model to be identified in Example \#2.}
\begin{tabular}{cc|c}
\hline
\multicolumn{2}{c|}{ Variable } & Description \\ \hline
\multicolumn{1}{c|}{\multirow{2}{*}{State $x$}}           & $x_{1}$ & Intake manifold pressure [$\unit[]{\kilo\pascal}$] \\ \cline{2-3} 
\multicolumn{1}{c|}{}                                     & $x_{2}$ & EGR rate, $\%$ \\ \hline
\multicolumn{1}{c|}{\multirow{4}{*}{Exogenous input $w$}} & $w_{1}$ & Engine speed [$\unit[]{\radian / \second}]$ \\ \cline{2-3} 
\multicolumn{1}{c|}{}                                     & $w_{2}$ & Fuel injection amount [$\unit[]{\milli\gram} / \mathrm{stroke}$] \\ \cline{2-3} 
\multicolumn{1}{c|}{}                                     & $w_{3}$ & EGR valve opening, \% \\ \cline{2-3} 
\multicolumn{1}{c|}{}                                     & $w_{4}$ & VGT vane closing, \%  \\ \hline
\end{tabular}
    \label{Table_Variables_of_model}
\end{table}


\subsection{Basis functions}  \label{sS_Example_Engine_Basis_functions}
In this example, we set the basis function as the constant and the polynomials shown in \eqref{Eq:theta_1_28} and the Gaussian radial basis functions (RBFs) shown in \eqref{Eq:theta_28+i}:
\begin{equation}
    \begin{gathered}
    \mathrm{\theta}_{1} = 1,  \;
    \mathrm{\theta}_{2} = x_{1},  \;
    \mathrm{\theta}_{3} = x_{2},  \;
    \mathrm{\theta}_{4} = w_{1},  \; \\
    \mathrm{\theta}_{5} = w_{2},  \;
    \mathrm{\theta}_{6} = w_{3},  \; 
    \mathrm{\theta}_{7} = w_{4},  \;
    \mathrm{\theta}_{8} = x_{1}^{2},  \; \\
    \mathrm{\theta}_{9} = x_{1}x_{2},  \;
    \mathrm{\theta}_{10} = x_{1}w_{1},  \;
    \mathrm{\theta}_{11} = x_{1}w_{2},  \; 
    \mathrm{\theta}_{12} = x_{1}w_{3},  \; \\
    \mathrm{\theta}_{13} = x_{1}w_{4},  \; 
    \mathrm{\theta}_{14} = x_{2}^{2},  \;
    \mathrm{\theta}_{15} = x_{2}w_{1},  \; 
    \mathrm{\theta}_{16} = x_{2}w_{2},  \; \\
    \mathrm{\theta}_{17} = x_{2}w_{3},  \; 
    \mathrm{\theta}_{18} = x_{2}w_{4},  \;
    \mathrm{\theta}_{19} = w_{1}^{2},  \; 
    \mathrm{\theta}_{20} = w_{1}w_{2},  \; \\
    \mathrm{\theta}_{21} = w_{1}w_{3},  \; 
    \mathrm{\theta}_{22} = w_{1}w_{4},  \;
    \mathrm{\theta}_{23} = w_{2}^{2},  \; 
    \mathrm{\theta}_{24} = w_{2}w_{3},  \; \\
    \mathrm{\theta}_{25} = w_{2}w_{4},  \; 
    \mathrm{\theta}_{26} = w_{3}^{2},  \;
    \mathrm{\theta}_{27} = w_{3}w_{4},  \;
    \mathrm{\theta}_{28} = w_{4}^{2}  \;
    \end{gathered}
    \label{Eq:theta_1_28}
\end{equation}
\begin{equation}
    \mathrm{\theta}_{28+i} = \mathrm{exp} \left\{ 
        -\left(u - \mu_{i}^{\top} \right)^{\top} \mathit{\Sigma}_{i}^{-2}  \left(u - \mu_{i}^{\top} \right) 
                                          \right\}
\label{Eq:theta_28+i}
\end{equation}
where
\begin{equation}
    u \triangleq \begin{bmatrix} x_{1} & x_{2} & w_{1} & w_{2} & w_{3} & w_{4} \end{bmatrix}^{\top}
    \label{Eq:u=[x,w]}
\end{equation}
\begin{equation}
    \mathit{\Sigma_{i}} \triangleq \mathrm{diag}\Of{\sigma_{i}}
    \label{Eq:Sigma=diag(sigma)}
\end{equation}
\begin{equation}
    \mu_{i} \triangleq \BvecS{\mu}{i,1}{i,2}{i,6}
    \label{Eq:mu=[mu_1_6]}
\end{equation}
\begin{equation}
    \sigma_{i} \triangleq \BvecS{\sigma}{i,1}{i,2}{i,6}
    \label{Eq:sigma=[sigma_1_6]}
\end{equation}
for $i = 1,\ldots,5$ and $\mu_{i,1}, \ldots, \mu_{i,5}, \sigma_{i,1}, \ldots, \sigma_{i,5} \in \mathbb{R}$. The tunable parameters for the Gaussian RBFs are aggregated as
\begin{equation}
    \varPhi = \begin{bmatrix} \mu_{1} & \cdots & \mu_{5} & \sigma_{1} & \cdots & \sigma_{5} \end{bmatrix}^{\top}.
    \label{Eq:Phi=[mu,sigma]}
\end{equation}
A linear combination of the Gaussian RBFs has the universal approximation property \cite{Park1991}. 
The Gaussian RBFs \eqref{Eq:theta_28+i} can be flexibly designed by adjusting the parameters $\mu_{i}$ and $\sigma_{i}$. 

\subsection{Verification setting}  \label{sS_Example_Engine_Verification_setting}
As in Section \ref{sS_Example_SLR_Verification_setting}, we compare SINDy-LOM with the conventional SINDy method. Table \ref{Table_Overview_of_modeling_strategies} summarizes the three different modeling strategies. Strategies \#1 and \#2 use the traditional SINDy method, where the libraries are fixed. For Strategy \#2, $\varPhi$ is randomly chosen from the interval of 
\textcolor{black}{$\left[-500, 500\right]^{60}$}
. The specific values of $\varPhi$ used for Strategy \#2, denoted as $\varPhi^{0}$, are listed in Table \ref{Table_Phi_Strategy_2}. Meanwhile, SINDy-LOM is employed for Strategy \#3, where $\varPhi$ is optimized according to the library optimization mechanism. The optimization problem \eqref{Eq:Phi^star=argmin(Jms)} was solved using GA (\texttt{ga} function implemented in MATLAB R2023a). The initial population is randomly selected from the interval of 
\textcolor{black}{$\left[-500, 500\right]^{60}$}
, and the search domain is set as whole of $\mathbb{R}^{60}$. The sparsification parameter $\lambda$ in the STLSQ algorithm is set to $8.0 \times 10^{-5}$ for Strategies \#1--\#3. For Strategy \#3, $\kappa = 8.0 \times 10^{-7}$ is used. 
In \eqref{Eq:Jms=def}, we set $q_{1} = q_{2} = 1$ and $r_{1} = r_{2} = 1$.
The implication of comparing Strategies \#1--\#3 is discussed in Remark \ref{Remark_Implication_of_comparison}.

\begin{figure}
    \begin{center}
    \includegraphics[width=8.65cm]{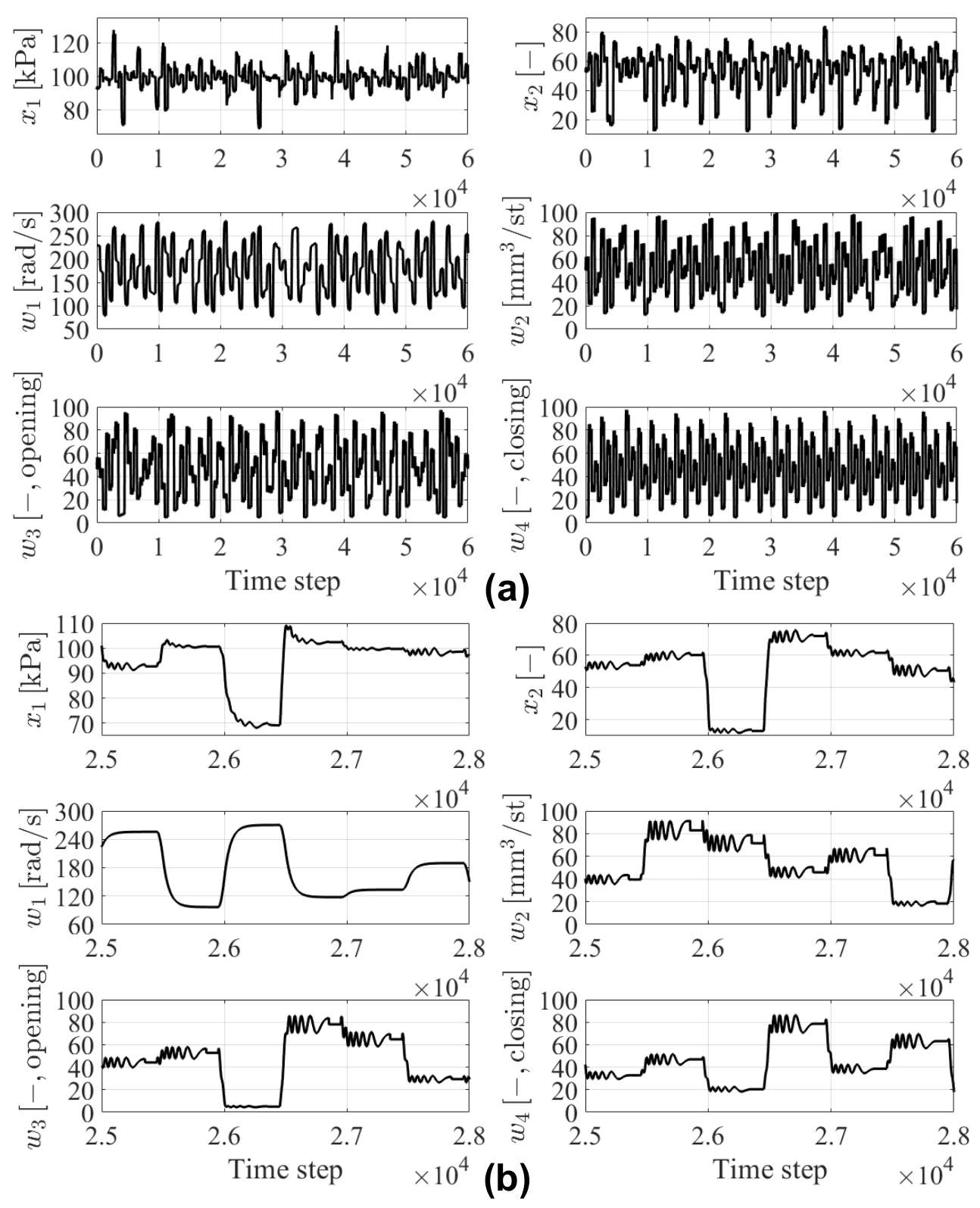}
    \caption{\protect{SR data (Example \#2): (a) overview, and (b) enlarged plot showing the typical behavior in detail. This data is used for both $\left(\mathcal{X}^{SR},\mathcal{W}^{SR} \right)$ and $\left(\mathcal{X}^{LL_{1}},\mathcal{W}^{LL_{1}} \right)$.}}
    \label{Fig_SR_Data}
    \end{center}
\end{figure}

\begin{figure}
    \begin{center}
    \includegraphics[width=8.65cm]{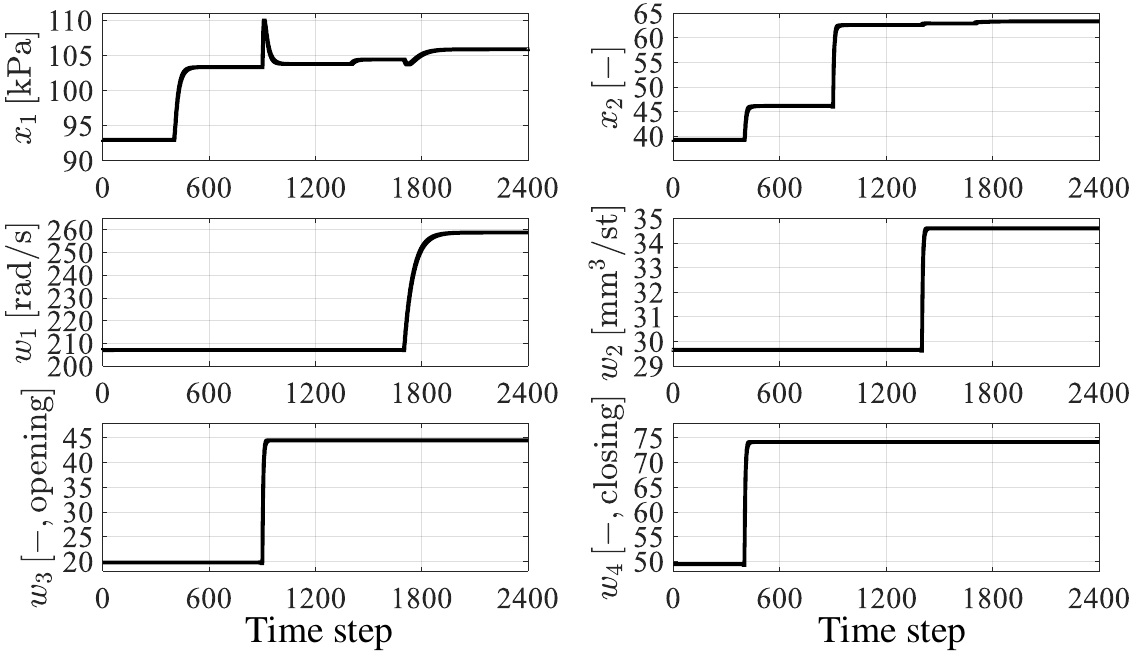}
    \caption{\protect{Overview of the oLL data for Example \#2. This data is used as $\left(\mathcal{X}^{LL_{2}},\mathcal{W}^{LL_{2}} \right)$.}}
    \label{Fig_oLL_Data}
    \end{center}
\end{figure}

The data used for system identification are shown in Figs. \ref{Fig_SR_Data} and \ref{Fig_oLL_Data}, with a sampling interval of $\qty[]{0.01}{\second}$.
Consistent with Example \#1, these datasets are referred to as the \emph{SR data} and the \emph{oLL data}, respectively. 
Strategies \#1--\#3 use the SR data for sparse regression. In Strategy \#3, the RLT prediction accuracy is evaluated using both the SR and oLL datasets.
In other words, we employ the SR data for both $\left(\mathcal{X}^{SR},\mathcal{W}^{SR} \right)$ and $\left(\mathcal{X}^{LL_{1}},\mathcal{W}^{LL_{1}} \right)$; the oLL data is used as $\left(\mathcal{X}^{LL_{2}},\mathcal{W}^{LL_{2}} \right)$. 
The number of data points is $N^{SR} = N_{1}^{LL} = 6.0 \times 10^{4}$ and $N_{2}^{LL} = 2.4 \times 10^{3}$.
Fig. \ref{Fig_oLL_Data} shows that the diesel engine airpath system has strong nonlinearity and coupling.

To obtain the SR data, the exogenous input signals $w_{1},\ldots,w_{4}$ are generated based on the Design of Experiments. Specifically, $w_{1},\ldots,w_{4}$ for the SR data are generated through the following steps: 1) the piecewise constant inputs are generated from the Sobol quasi-random sequences; 2) down-chirp fluctuations are applied to the piecewise inputs corresponding to $w_{2},\ldots,w_{4}$; 3) these piecewise signals are then passed through the low-pass filters; 4) the filtered signals are used as $w_{1},\ldots,w_{4}$. The oLL data were collected by setting the low-pass-filtered step signals as $w_{1},\ldots,w_{4}$ and applying them to the system.
The motivation behind selecting such SR data and oLL data is the same as that used in Example \#1.

\begin{table}[]
    \centering
    \caption{Overview of modeling strategies for Example \#2.}
    {
        \tabcolsep = 1.5pt   
\begin{tabular}{c|c|c|c}
\hline
Strategy & Algorithm & Library & $\varPhi$ \\ \hline
\#1 & \begin{tabular}{c} SINDy \\ (Conventional) \end{tabular} & \begin{tabular}{c}$\ThetaOf{1}{x\Of{k},w\Of{k}}$ \\ $=\BvecS{\mathrm{\theta}}{1}{2}{28}$\end{tabular} & -- \\ \hline
\#2 & \begin{tabular}{c} SINDy \\ (Conventional) \end{tabular} & \begin{tabular}{c}$\ThetaOf{2}{x\Of{k},w\Of{k}; \varPhi}$ \\ $=\BvecS{\mathrm{\theta}}{1}{2}{33}$\end{tabular} & \begin{tabular}{c} Fixed, \\ randomly chosen \end{tabular} \\ \hline
\#3 & \begin{tabular}{c} SINDy-LOM \\ (Proposed) \end{tabular} & \begin{tabular}{c}$\ThetaOf{3}{x\Of{k},w\Of{k}; \varPhi}$ \\ $=\BvecS{\mathrm{\theta}}{1}{2}{33}$\end{tabular} & \begin{tabular}{c} Optimized during \\ the modeling process \end{tabular} \\ \hline
\end{tabular}
    }
\label{Table_Overview_of_modeling_strategies}
\end{table}


\begin{table}[]
    \centering
    \caption{Design variable $\varPhi^{0}$ for the Gaussian RBFs in the library for Strategy \#2 (Example \#2).}
    {\renewcommand\arraystretch{1.1}   
\begin{tabular}{c|c|c}
\hline
\multirow{2}{*}{$\mathit{\theta}_{29}$} & $\mu_{1}$    & \begin{tabular}{c}
{\renewcommand\arraystretch{0.9} 
    $
\left[
\begin{array}{ccc}
-82.9780 & 220.3245 & -499.8856 
\end{array}
\right.
$
}
\\
{\renewcommand\arraystretch{0.9}  
$
\left.
\begin{array}{ccc}
-197.6674 & -353.2441 & -407.6614 
\end{array}
\right]
$
}
\end{tabular}

 \\ \cline{2-3} 
                                        & $\sigma_{1}$ & \begin{tabular}{c}
    {\renewcommand\arraystretch{0.9}  
    $
\left[
\begin{array}{ccc}
-401.6532 & -78.8924 & 457.8895
\end{array}
\right.
$
    }
\\
{\renewcommand\arraystretch{0.9} 
$
\left.
\begin{array}{ccc}
33.1653 & 191.8771 & -184.4844
\end{array}
\right]
$
}
\end{tabular}
 \\ \hline
\multirow{2}{*}{$\mathit{\theta}_{30}$} & $\mu_{2}$    & \begin{tabular}{c}
{\renewcommand\arraystretch{0.9}  
    $
\left[
\begin{array}{ccc}
-313.7398 & -154.4393 & -103.2325
\end{array}
\right.
$
}
\\
{\renewcommand\arraystretch{0.9}  
$
\left.
\begin{array}{ccc}
38.8167 & -80.8055 & 185.2195
\end{array}
\right]
$
}
\end{tabular}
 \\ \cline{2-3} 
                                        & $\sigma_{2}$ & \begin{tabular}{c}
    {\renewcommand\arraystretch{0.9}  
    $
\left[
\begin{array}{ccc}
186.5009 & 334.6257 & -481.7117
\end{array}
\right.
$
    }
\\
{\renewcommand\arraystretch{0.9}  
$
\left.
\begin{array}{ccc}
250.1443 & 488.8611 & 248.1657
\end{array}
\right]
$
}
\end{tabular}

 \\ \hline
\multirow{2}{*}{$\mathit{\theta}_{31}$} & $\mu_{3}$    & \begin{tabular}{c}
{\renewcommand\arraystretch{0.9}  
    $
\left[
\begin{array}{ccc}
-295.5478 & 378.1174 & -472.6124
\end{array}
\right.
$
}
\\
{\renewcommand\arraystretch{0.9}  
$
\left.
\begin{array}{ccc}
170.4675 & -82.6952 & 58.6898
\end{array}
\right]
$
}
\end{tabular}
 \\ \cline{2-3} 
                                        & $\sigma_{3}$ & \begin{tabular}{c}
    {\renewcommand\arraystretch{0.9}  
    $
\left[
\begin{array}{ccc}
-219.5560 & 289.2793 & -396.7740
\end{array}
\right.
$
    }
\\
{\renewcommand\arraystretch{0.9}  
$
\left.
\begin{array}{ccc}
-52.1065 & 408.5955 & -206.3859
\end{array}
\right]
$
}
\end{tabular}
 \\ \hline
\multirow{2}{*}{$\mathit{\theta}_{32}$} & $\mu_{4}$    & \begin{tabular}{c}
{\renewcommand\arraystretch{0.9}  
    $
\left[
\begin{array}{ccc}
-359.6131 & -301.8985 & 300.7446
\end{array}
\right.
$
}
\\
{\renewcommand\arraystretch{0.9}  
$
\left.
\begin{array}{ccc}
468.2616 & -186.5758 & 192.3226
\end{array}
\right]
$
}
\end{tabular}
 \\ \cline{2-3} 
                                        & $\sigma_{4}$ & \begin{tabular}{c}
    {\renewcommand\arraystretch{0.9} 
    $
\left[
\begin{array}{ccc}
-212.2247 & -369.9714 & -480.6330
\end{array}
\right.
$
    }
\\
{\renewcommand\arraystretch{0.9}  
$
\left.
\begin{array}{ccc}
178.8355 & -288.3719 & -234.4533
\end{array}
\right]
$
}
\end{tabular}
 \\ \hline
\multirow{2}{*}{$\mathit{\theta}_{33}$} & $\mu_{5}$    & \begin{tabular}{c}
{\renewcommand\arraystretch{0.9}  
    $
\left[
\begin{array}{ccc}
376.3892 & 394.6067 & -414.9558
\end{array}
\right.
$
}
\\
{\renewcommand\arraystretch{0.9}  
$
\left.
\begin{array}{ccc}
-460.9452 & -330.1696 & 378.1425
\end{array}
\right]
$
}
\end{tabular}

 \\ \cline{2-3} 
                                        & $\sigma_{5}$ & \begin{tabular}{c}
    {\renewcommand\arraystretch{0.9}  
    $
\left[
\begin{array}{ccc}
-8.4268 & -446.6375 & 74.1176
\end{array}
\right.
$
    }
\\
{\renewcommand\arraystretch{0.9}  
$
\left.
\begin{array}{ccc}
-353.2714 & 89.3055 & 199.7584
\end{array}
\right]
$
}
\end{tabular}

 \\ \hline
\end{tabular}
    }
    \label{Table_Phi_Strategy_2}
\end{table}


\subsection{Results}  \label{sS_Example_Engine_Results}
\begin{figure}
    \begin{center}
    \includegraphics[width=9.0cm]{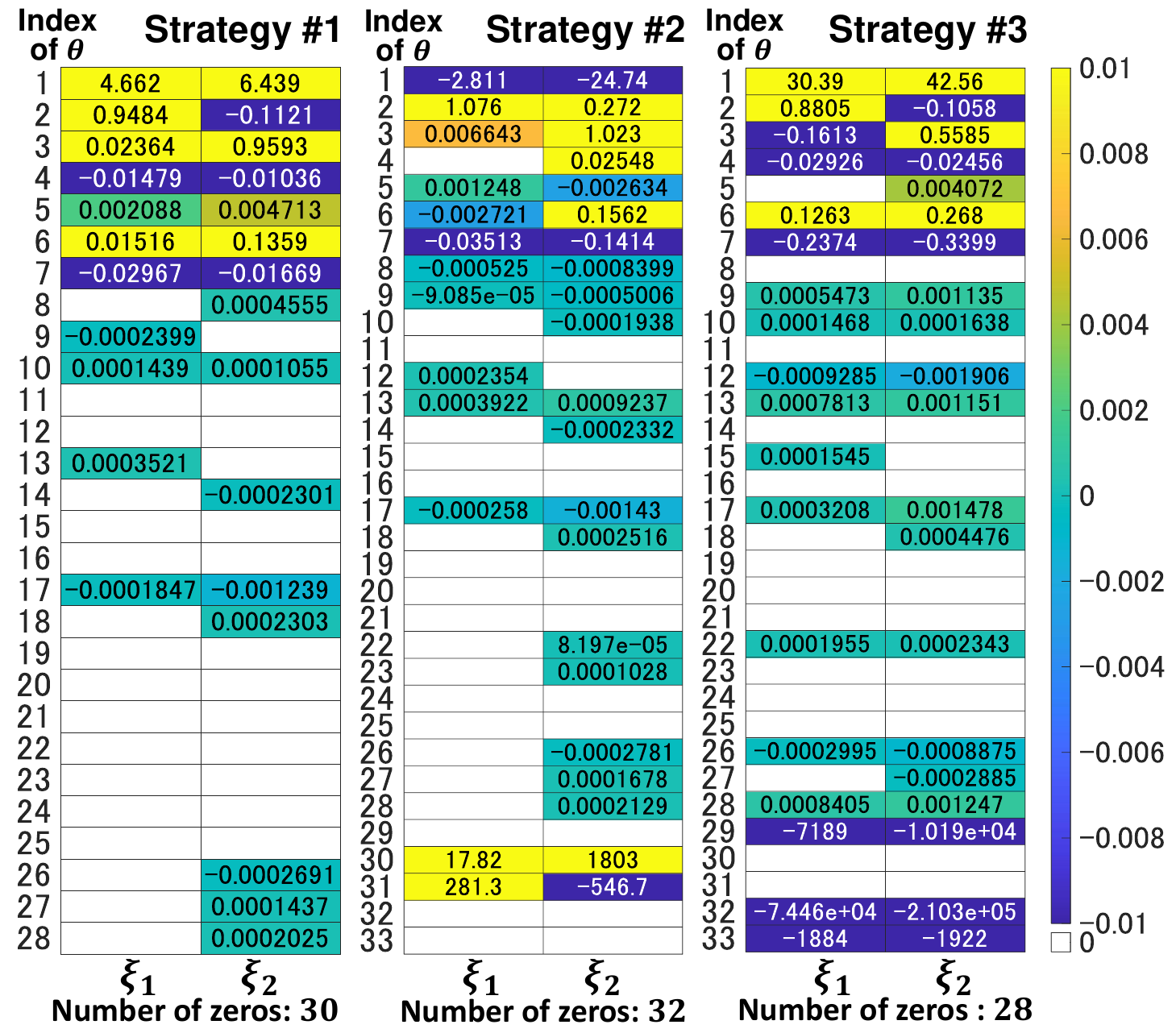}
    \caption{\protect{Visualization of the resulting coefficient matrix $\varXi^{\ast} = \begin{bmatrix} \xi_{1}^{\ast} & \xi_{2}^{\ast} \end{bmatrix}$ obtained by Strategies \#1--\#3 (Example \#2).}}
    \label{Fig_Xi_Coefficients}
    \end{center}
\end{figure}

Fig. \ref{Fig_Xi_Coefficients} shows the coefficient matrix $\varXi^{\ast} = \begin{bmatrix} \xi_{1}^{\ast} & \xi_{2}^{\ast} \end{bmatrix}$ obtained by Strategies \#1--\#3. In Fig. \ref{Fig_Xi_Coefficients}, each raw index indicates the corresponding basis function; the white box implies that the element is zero. 
Fig. \ref{Fig_Xi_Coefficients} shows that Strategies \#1--\#3 provide the parsimonious data-driven models, as all the $\varXi^{\ast}$ contain many zero elements. The design variable $\varPhi$ of the library of Strategy \#3, denoted by $\varPhi^{\ast}$, is listed in Table \ref{Table_Phi_Strategy_3}. 

\begin{figure}
    \begin{center}
    \includegraphics[width=8.95cm]{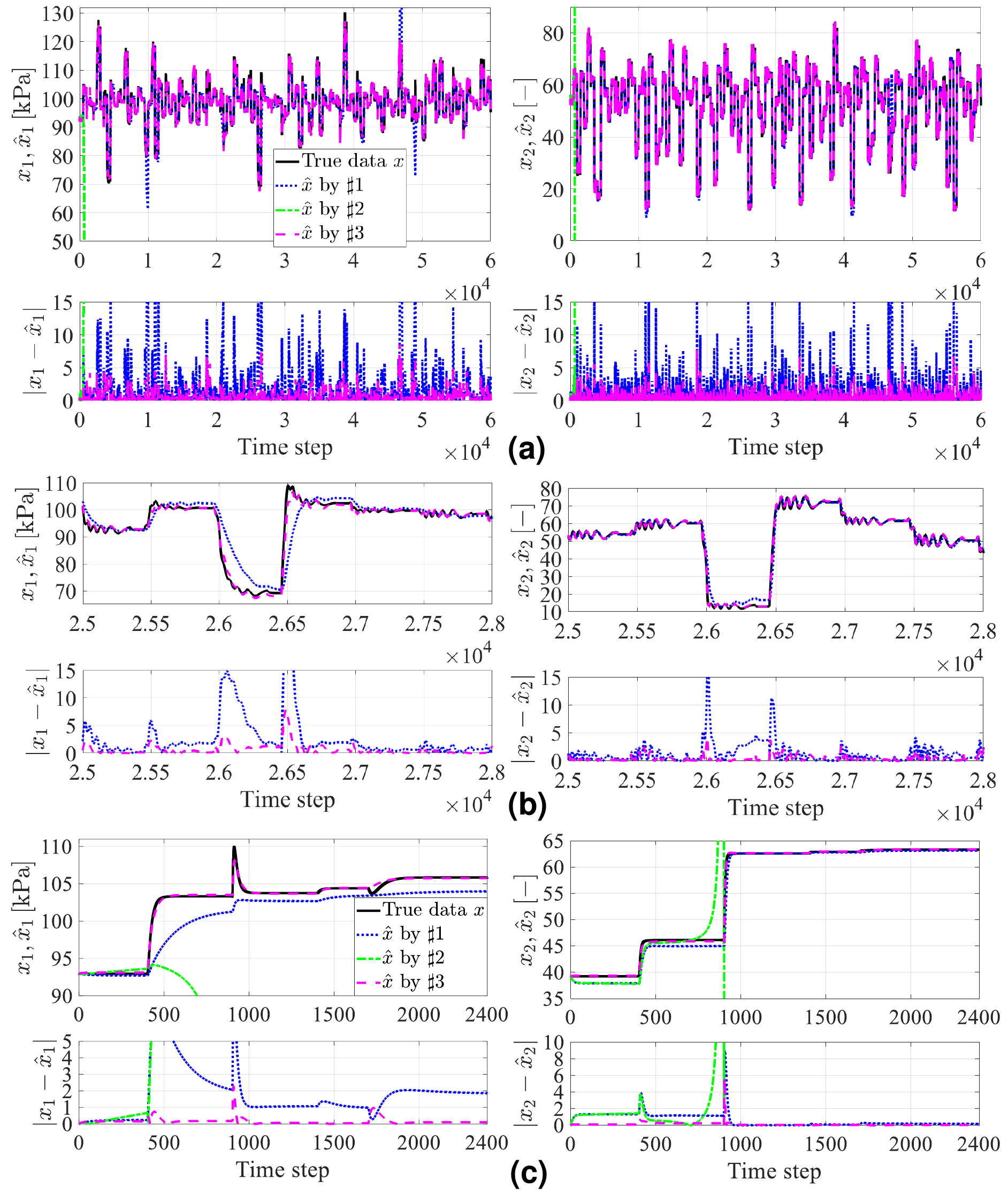}
    \caption{\protect\input{Fig_RLT_SR_oLL_Caption.tex}}
    \label{Fig_RLT_SR_oLL}
    \end{center}
\end{figure}

In the same manner as Example \#1, let $x\Of{k}$ and $\hat{x}\Of{k}$ denote the state of the true data and its RLT prediction, respectively. Note that the RLT prediction is produced in the same manner as \eqref{Eq:xhat^LLi(k+1)} and \eqref{Eq:xhat^LLi(0)}. 
Fig. \ref{Fig_RLT_SR_oLL} show the RLT predictions for the SR data and oLL data. 
The meaning of each line color in Fig. \ref{Fig_RLT_SR_oLL} is the same as in Fig. \ref{Fig_SLR_RLT_SR_oLL}. Table \ref{Table_2_Norm_RLT} summarizes the RLT prediction errors. 
Fig. \ref{Fig_RLT_SR_oLL} and Table \ref{Table_2_Norm_RLT} clearly show that Strategy \#3 achieved a much more accurate RLT prediction than Strategies \#1 and \#2. 
Moreover, the one-step-ahead prediction results obtained by Strategies \#1--\#3 are presented in Appendix \ref{sS_Appendix_One_step_prediction}. A comparison between Figs. \ref{Fig_RLT_SR_oLL} and \ref{Fig_One_step_SR}(b) reveals the discrepancy between the one-step-ahead and RLT predictions.

\begin{table}[]
    \centering
    \caption{Design variable $\varPhi^{\ast}$ for the Gaussian RBFs in the library for Strategy \#3 (Example \#2). These values are optimized by the proposed library optimization mechanism.}
    {\renewcommand\arraystretch{1.1}   
    \tabcolsep = 3.0pt   
\begin{tabular}{c|c|c}
\hline
\multirow{2}{*}{$\mathit{\theta}_{29}$} & $\mu_{1}$    & \begin{tabular}{c}
    {\renewcommand\arraystretch{0.9}  
    $
\left[
\begin{array}{ccc}
-8.2370\times10^{2} & 9.2166\times10^{1} & -2.8113\times10^{2}
\end{array}
\right.
$
    }
\\
{\renewcommand\arraystretch{0.9}  
$
\left.
\begin{array}{ccc}
-2.7292\times10^{2} & -1.1298\times10^{3} & -9.0266\times10^{2}
\end{array}
\right]
$
}
\end{tabular}

 \\ \cline{2-3} 
                                        & $\sigma_{1}$ & \begin{tabular}{c}
    {\renewcommand\arraystretch{0.9} 
    $
\left[
\begin{array}{ccc}
-1.7763\times10^{3} & -3.1180\times10^{2} & 1.3449\times10^{3}
\end{array}
\right.
$
    }
\\
{\renewcommand\arraystretch{0.9} 
$
\left.
\begin{array}{ccc}
6.8238\times10^{3} & -2.0355\times10^{3} & 3.8125\times10^{2}
\end{array}
\right]
$
}
\end{tabular}
 \\ \hline
\multirow{2}{*}{$\mathit{\theta}_{30}$} & $\mu_{2}$    & \begin{tabular}{c}
    {\renewcommand\arraystretch{0.9}  
    $
\left[
\begin{array}{ccc}
-1.0499\times10^{3} & 1.3560\times10^{3} & -2.1153\times10^{3}
\end{array}
\right.
$
    }
\\
{\renewcommand\arraystretch{0.9} 
$
\left.
\begin{array}{ccc}
4.0910 \times 10^{2} & -1.8164\times10^{2} & -1.4763\times10^{3}
\end{array}
\right]
$
}
\end{tabular}

 \\ \cline{2-3} 
                                        & $\sigma_{2}$ & \begin{tabular}{c}
    {\renewcommand\arraystretch{0.9}  
    $
\left[
\begin{array}{ccc}
-2.9288\times10^{2} & 3.5331\times10^{2} & 2.1096\times10^{3}
\end{array}
\right.
$
    }
\\
{\renewcommand\arraystretch{0.9} 
$
\left.
\begin{array}{ccc}
1.3085\times10^{3} & 2.4203\times10^{3} & 4.0081\times10^{3}
\end{array}
\right]
$
}
\end{tabular}
 \\ \hline
\multirow{2}{*}{$\mathit{\theta}_{31}$} & $\mu_{3}$    & \begin{tabular}{c}
    {\renewcommand\arraystretch{0.9}  
    $
\left[
\begin{array}{ccc}
-7.5463\times10^{2} & -1.9000\times10^{3} & 9.6181\times10^{2}
\end{array}
\right.
$
    }
\\
{\renewcommand\arraystretch{0.9} 
$
\left.
\begin{array}{ccc}
-8.1925\times10^{1} & -2.1613\times10^{2} & -1.3807\times10^{3}
\end{array}
\right]
$
}
\end{tabular}

 \\ \cline{2-3} 
                                        & $\sigma_{3}$ & \begin{tabular}{c}
    {\renewcommand\arraystretch{0.9}  
    $
\left[
\begin{array}{ccc}
1.5366 \times 10^{2} & -4.5731\times10^{3} & 2.2707\times10^{3}
\end{array}
\right.
$
    }
\\
{\renewcommand\arraystretch{0.9} 
$
\left.
\begin{array}{ccc}
-5.8682\times10^{2} & -1.4369\times10^{3} & 1.5524\times10^{3}
\end{array}
\right]
$
}
\end{tabular}
 \\ \hline
\multirow{2}{*}{$\mathit{\theta}_{32}$} & $\mu_{4}$    & \begin{tabular}{c}
    {\renewcommand\arraystretch{0.9}  
    $
\left[
\begin{array}{ccc}
2.2715\times10^{3} & -1.6417\times10^{3} & 2.2714\times10^{3}
\end{array}
\right.
$
    }
\\
{\renewcommand\arraystretch{0.9} 
$
\left.
\begin{array}{ccc}
5.2269\times10^{1} & -2.2746\times10^{2} & 9.2615\times10^{2}
\end{array}
\right]
$
}
\end{tabular}

 \\ \cline{2-3} 
                                        & $\sigma_{4}$ & \begin{tabular}{c}
    {\renewcommand\arraystretch{0.9}  
    $
\left[
\begin{array}{ccc}
-2.9782\times10^{2} & 3.4149\times10^{3} & 3.0367\times10^{3}
\end{array}
\right.
$
    }
\\
{\renewcommand\arraystretch{0.9}  
$
\left.
\begin{array}{ccc}
-2.5931\times10^{2} & 8.0149\times10^{1} & -1.7126\times10^{3}
\end{array}
\right]
$
}
\end{tabular}
 \\ \hline
\multirow{2}{*}{$\mathit{\theta}_{33}$} & $\mu_{5}$    & \begin{tabular}{c}
    {\renewcommand\arraystretch{0.9}  
    $
\left[
\begin{array}{ccc}
-1.3943\times10^{3} & 3.5249\times10^{2} & 7.6109\times10^{2}
\end{array}
\right.
$
    }
\\
{\renewcommand\arraystretch{0.9}  
$
\left.
\begin{array}{ccc}
-2.6745\times10^{2} & -1.8572\times10^{3} & 2.7690\times10^2
\end{array}
\right]
$
}
\end{tabular}

 \\ \cline{2-3} 
                                        & $\sigma_{5}$ & \begin{tabular}{c}
    {\renewcommand\arraystretch{0.9}  
    $
\left[
\begin{array}{ccc}
-1.2819\times10^{3} & -1.9387\times10^{2} & -5.6673\times10^{2}
\end{array}
\right.
$
    }
\\
{\renewcommand\arraystretch{0.9} 
$
\left.
\begin{array}{ccc}
-4.1669\times10^{2} & -2.5020\times10^{3} & -2.3898\times10^{2}
\end{array}
\right]
$
}
\end{tabular}

 \\ \hline
\end{tabular}
    }
    \label{Table_Phi_Strategy_3}
\end{table}


\begin{table}[]
    \centering
    \caption{Errors of the RLT predictions (Example \#2).}
    {\renewcommand\arraystretch{1.1}   
\begin{tabular}{c|cccc}
\hline
\multirow{4}{*}{Strategy} & \multicolumn{4}{c}{\multirow{2}{*}{\raisebox{8pt}{Error of RLT prediction}}}                                   \\
                  & \multicolumn{4}{c}{$
    10^{2} \times \sqrt{\sum_{k} \left\lvert \hat{x}_{i}\Of{k} - x_{i}\Of{k}\right\rvert^{2}} 
    /
    \sqrt{\sum_{k} \left\lvert x_{i}\Of{k}\right\rvert^{2}}
$}                          \\ \cline{2-5} 
                  & \multicolumn{2}{c|}{SR}                       & \multicolumn{2}{c}{oLL}    \\ \cline{2-5} 
                  & \multicolumn{1}{c|}{\phantom{xxx}$x_{1}$\phantom{xxx}} & \multicolumn{1}{c|}{\phantom{xxx}$x_{2}$\phantom{xxx}} & \multicolumn{1}{c|}{\phantom{xxx}$x_{1}$\phantom{xxx}} & \multicolumn{1}{c}{\phantom{xxx}$x_{2}$\phantom{xxx}}  \\ \hline
\#1               & \multicolumn{1}{c|}{$4.1754$} & \multicolumn{1}{c|}{$4.7299$} & \multicolumn{1}{c|}{$2.3020$} & \multicolumn{1}{c}{$1.8604$}  \\ \hline
\#2               & \multicolumn{1}{c|}{diverged} & \multicolumn{1}{c|}{diverged} & \multicolumn{1}{c|}{diverged} & \multicolumn{1}{c}{diverged}  \\ \hline
\#3               & \multicolumn{1}{c|}{$1.1453$} & \multicolumn{1}{c|}{$1.3361$} & \multicolumn{1}{c|}{$0.25574$} & \multicolumn{1}{c}{$0.49183$}  \\ \hline
\end{tabular}
    }
    \label{Table_2_Norm_RLT}
\end{table}


\section{Discussion}  \label{S_Discussion}
The numerical examples presented in Sections \ref{S_Example_SLR_Numerical_example} and \ref{S_Example_Engine_Numerical_example} confirm the validity of the proposed approach.
In both examples, the models obtained by Strategy \#1 yield poor RLT predictions, as demonstrated by Fig. \ref{Fig_SLR_RLT_SR_oLL} (Example \#1) and Fig. \ref{Fig_RLT_SR_oLL} along with Table \ref{Table_2_Norm_RLT} (Example \#2).
Although Strategy \#2 employs a richer library than Strategy \#1, its RLT prediction performance shows no improvement.
A comparison between Strategies \#1 and \#2 highlights the challenge of library design in the SINDy framework.
Meanwhile, the data-driven model obtained by Strategy \#3, namely, the proposed approach, achieves the most accurate RLT predictions.
This result arises from the proposed library optimization mechanism.
The only difference between Strategies \#3 and \#2 lies in whether the parameter $\varPhi$ of the basis functions is optimized.
The proposed mechanism can search for appropriate basis functions based on an optimization algorithm, thereby reducing the user's burden.
Therefore, the proposed approach provides a powerful strategy to address the problem of library design in the SINDy framework.
The proposed approach is particularly suitable for modeling dynamical systems where the appropriate basis functions and their parameters cannot be exactly determined \emph{a priori}. Most industrial systems fall into this category.

In Example \#1, Fig. \ref{Fig_SLR_Noisy_RLT_SR_oLL} and Table \ref{Table_SLR_Noisy_Result} demonstrate that the proposed approach is robust to data noise. As shown in Table \ref{Table_SLR_Noisy_Result}, it can accurately recover the ground-truth dynamics even when the training data are corrupted by noise. Fig. \ref{Fig_SLR_Noisy_RLT_SR_oLL} illustrates that the resulting models yield accurate RLT predictions of the ground-truth system trajectories. 
However, the identification performance of SINDy-LOM gradually deteriorated with increasing noise intensity; for example, at $\sigma_{x} = 0.5$, the equation for $x_{1}\Of{k+1}$ contains an erroneous term. 
Nevertheless, the degradation was not critical, as the RLT prediction successfully captured the dynamical behavior of the ground-truth system.
Future work will focus on enhancing the noise robustness of SINDy-LOM by incorporating advanced learning strategies, such as ensemble methods and other sparsification techniques based on criteria other than coefficient thresholding.   

Comparing RLT and one-step-ahead prediction performance highlights the importance of RLT prediction. 
In Examples \#1 and \#2, the one-step-ahead prediction accuracies of Strategies \#1 and \#2 are comparable to those of Strategy \#3 (Fig. \ref{Fig_One_step_SR}).
However, they fail to provide accurate RLT predictions throughout the prediction horizon (Figs. \ref{Fig_SLR_RLT_SR_oLL} and \ref{Fig_RLT_SR_oLL}). This inconsistency underscores the importance of explicitly considering RLT prediction. The proposed approach explicitly accounts for RLT prediction, thereby ensuring consistency with the true system and enhancing model reliability.

Example \#2 demonstrates that the proposed approach provides an effective data-driven modeling technique for industrial applications. 
As illustrated in Fig. \ref{Fig_oLL_Data}, the diesel engine airpath system exhibits strong nonlinearity and coupling. Despite this complexity, the proposed approach successfully identifies a model that provides accurate RLT predictions.
Moreover, the model obtained by the proposed approach is interpretable, providing a parsimonious closed-form expression (Fig. \ref{Fig_Xi_Coefficients}). This interpretability is important for understanding the model and the underlying dynamical system, and it plays a critical role in adopting data-driven modeling techniques for engineering applications. Future work will apply the SINDy-LOM framework to other industrial systems.

This study addressed an important challenge in the SINDy framework, namely, the library design. Nevertheless, some issues remain in terms of enhancing its usability. One issue is the selection of the parametrization of the library: we must specify the parametrization of the basis functions. Although the selection of the class of the basis functions is easier than determining the functions and their design parameters, the design of the parametrized basis functions must involve some trial and error. Moreover, the optimization problem \eqref{Eq:Phi^star=argmin(Jms)} provides an important research direction. Specifically, \eqref{Eq:Phi^star=argmin(Jms)} is a non-convex optimization problem. Although \eqref{Eq:Phi^star=argmin(Jms)} can be handled by a standard numerical software tool in practice as demonstrated in this study, the convex reformulation of \eqref{Eq:Phi^star=argmin(Jms)} should reduce the computational complexity of the proposed approach.  In the future, we will tackle these issues to enhance the practical usability of the SINDy-LOM approach.


\section{Conclusion}  \label{S_Conclusion}
We developed a novel data-driven modeling approach for nonlinear dynamical systems. The proposed approach employs the SINDy framework. In contrast to the conventional SINDy technique, the proposed technique automatically designs the basis functions. We reformulated the library design procedure in the SINDy framework into the optimization of the parametrized basis functions. 
The data-driven modeling procedure has a two-layer optimization architecture. In the inner-layer, solving the sparse regression problem distills the data-driven model as a parsimonious linear combination of the candidate basis functions. In the outer-layer, the parametrized basis functions are modified to improve the RLT prediction accuracy of the model distilled in the inner-layer. By solving this two-layer optimization problem, the proposed approach discovers a parsimonious data-driven model that achieves high RLT prediction accuracy. 
The numerical example inspired by the single-link robot system demonstrates that the proposed approach successfully discovered the dynamical model that is almost the same as the ground-truth system in the presence of the data noise. 
In the numerical example involving the diesel engine airpath system, a challenging industrial application, the data-driven model identified by the proposed approach accurately captured the complex dynamic behavior of the system throughout the prediction horizon, whereas the conventional SINDy method failed to do so. Moreover, these numerical studies revealed the challenge of designing an appropriate library and the importance of incorporating RLT prediction accuracy into the SINDy framework. 
Consequently, this study offers a reliable and practical data-driven modeling framework.

In the future, we will use the proposed approach for constructing the prediction model in MPC.


\section*{Acknowledgments}
The authors thank Dr. Hiroki Seto of the ISUZU Advanced Engineering Center for fruitful discussion and suggestion.

\appendix
\section*{Inconsistency of the one-step-ahead and recursive long-term predictions}
\begin{figure}
    \begin{center}
    \includegraphics[width=8.95cm]{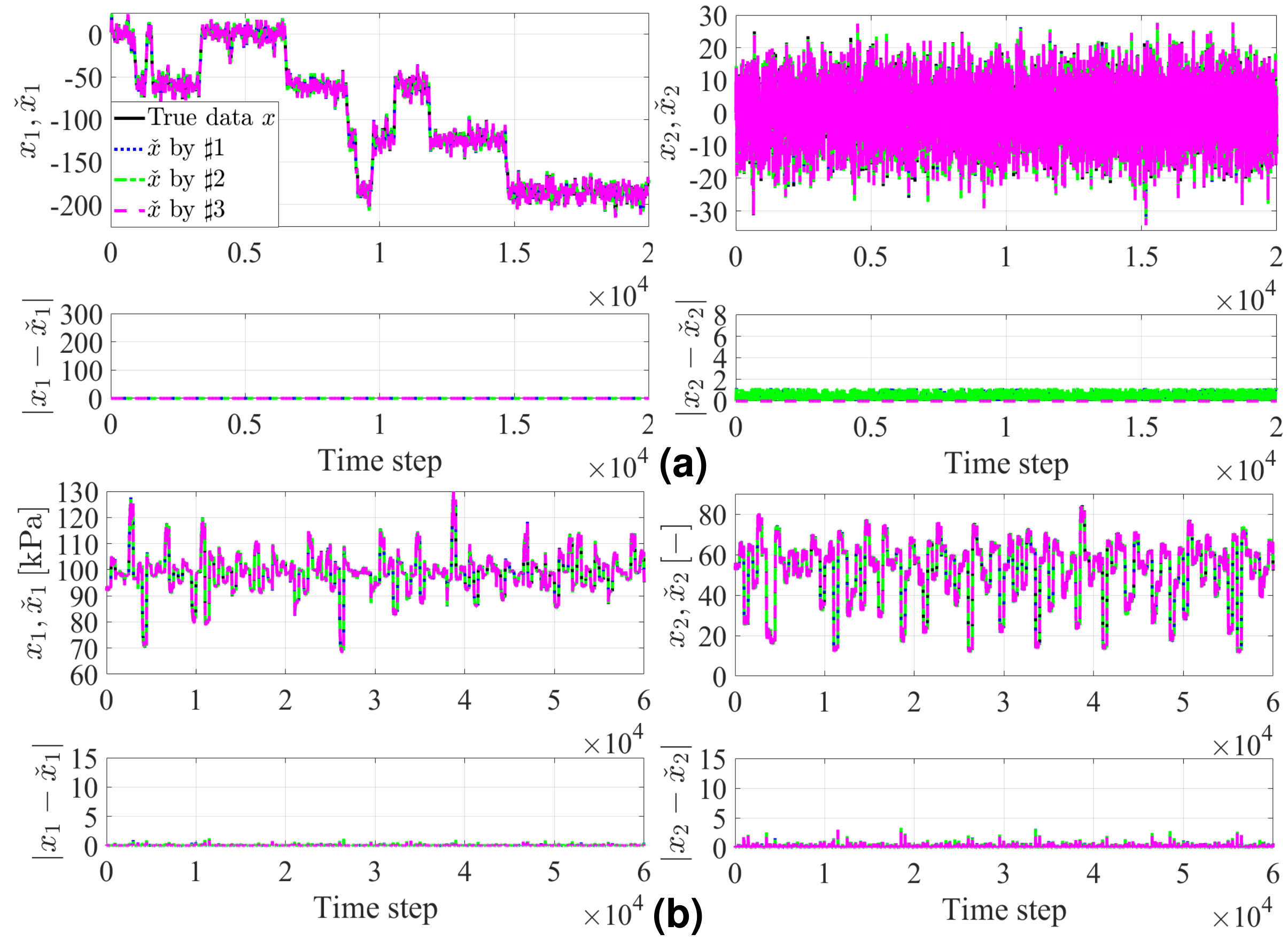}
    \caption{\protect\input{Fig_One_step_SR_Caption.tex}}
    \label{Fig_One_step_SR}
    \end{center}
\end{figure}

This appendix presents the one-step-ahead prediction results obtained by Strategies \#1--\#3. As discussed in Section \ref{sS_Challenges_in_SINDy}, the sparse least-squares regression in the conventional SINDy approach can be regarded as the minimization of the error between the true data $x^{D}\Of{k}$ and its one-step-ahead prediction $\check{x}^{D}\Of{k}$. Fig. \ref{Fig_One_step_SR} shows the one-step-ahead prediction results in Examples \#1 and \#2, in the same manner as \eqref{Eq:x_Check(k+1)=Theta(xk,wk)Xi}, provided by Strategies \#1--\#3. The one-step-ahead predictions are executed for the SR data. In contrast to the RLT predictions in Figs. \ref{Fig_SLR_RLT_SR_oLL} and \ref{Fig_RLT_SR_oLL}, Strategies \#1--\#3 provide comparable prediction accuracies in the one-step-ahead case. The inconsistency between the one-step-ahead and RLT predictions underscores the necessity of considering the RLT prediction.  

  \label{sS_Appendix_One_step_prediction}

\bibliographystyle{IEEEtran}        
\bibliography{References_IEEE}           


\end{document}